\documentclass[twoside,11pt]{article}
\usepackage{amsmath}
\usepackage{bm}
\usepackage{jmlr2e}
\usepackage{algorithm}
\usepackage{algorithmic}
\usepackage{graphicx}
\usepackage{epstopdf}
\usepackage{wrapfig}
\usepackage{subfigure}
\usepackage[utf8]{inputenc} 
\usepackage[T1]{fontenc}    
\usepackage{hyperref}       
\usepackage{url}            
\usepackage{booktabs}       
\usepackage{amsfonts}       
\usepackage{nicefrac}       
\usepackage{microtype}      

\jmlrheading{1}{2019}{1-25}{0/00}{00/00}{Minghao Gu and Shiliang Sun}

\ShortHeadings{Variational Langevin Hamiltonian Monte Carlo for Distant Multi-modal Sampling}{Minghao Gu and Shiliang Sun}
\firstpageno{1}

\begin{document}

\title{Variational Langevin Hamiltonian Monte Carlo for Distant Multi-modal Sampling}

\author{\name Minghao Gu \email guminghao1081@gmail.com \\
       \addr Department of Computer Science and Technology\\
       East China Normal University\\
       3663 North Zhongshan Road, Shanghai 200241, P. R. China
       \AND
       \name Shiliang Sun \email slsun@cs.ecnu.edu.cn \\
       \addr Department of Computer Science and Technology\\
       East China Normal University\\
       3663 North Zhongshan Road, Shanghai 200241, P. R. China}

\editor{}

\maketitle

\begin{abstract}
The Hamiltonian Monte Carlo (HMC) sampling algorithm exploits Hamiltonian dynamics to construct efficient Markov Chain Monte Carlo (MCMC), which has become increasingly popular in machine learning and statistics. Since HMC uses the gradient information of the target distribution, it can explore the state space much more efficiently than the random-walk proposals. However, probabilistic inference involving multi-modal distributions is very difficult for standard HMC method, especially when the modes are far away from each other. Sampling algorithms are then often incapable of traveling across the places of low probability. In this paper, we propose a novel MCMC algorithm which aims to sample from multi-modal distributions effectively. The method improves Hamiltonian dynamics to reduce the autocorrelation of the samples and uses a variational distribution to explore the phase space and find new modes. A formal proof is provided which shows that the proposed method can converge to target distributions. Both synthetic and real datasets are used to evaluate its properties and performance. The experimental results verify the theory and show superior performance in multi-modal sampling.
\end{abstract}


\section{Introduction}
Generating samples from the multiple distributions encountered in Bayesian inference and machine learning is difficult. Markov chain Monte Carlo (MCMC) is used to be a robust framework to generate the samples from the complex target distributions. Through constructing the specific Markov chains, the MCMC methods can efficiently converge to the correct target distribution with the chains evolving. Nowadays, MCMC plays an essential role in artificial intelligence applications and probability inference, especially for estimating the expectations of the target functions.

Sampling methods based on dynamics is one of the most popular MCMC methods. The most commonly used dynamics in MCMC are Langevin dynamics and Hamiltonian dynamics. Hamiltonian Monte Carlo (HMC) \citep{duane1987hybrid,Neal2010} has become one of the most popular MCMC algorithms in Bayesian inference and machine learning. Unlike the previous MCMC algorithms \citep{neal2003slice}, HMC takes advantage of the gradient information to explore the continuous probability density function (PDF), which makes HMC more efficient to converge to the target distribution. Mainly, HMC tansforms the PDF into the potential energy function and adds the kinetic energy function to simulate the motion of the particle in the particular phase space, and thus HMC is able to satisfy the ergodic property. In practice, HMC exploits the Hamiltonian equation to calculate the new state of the proposed points in the phase space. In order to keep the detailed balance, the Metropolis-Hasting technique is widely adopted \citep{martino2013flexibility}. Since gradient information helps to discover and explore the phase space more efficiently, HMC has much further research \citep{wang2013adaptive,hoffman2014no}.

Nevertheless, samplers based on dynamics still have some deficiencies. The traditional dynamics samplers \citep{neal1993probabilistic} and their deep research \citep{celeux2000computational,neal2001annealed,rudoy2006monte,Girolami2011, hoffman2014no} have excellent performance in unimodal distributions. However, when facing multi-modal distribution, these algorithms may meet some problems, especially when the modes are far away from each other. When the modes are close to each other, the momentum variable in dynamics samplers may offer chances for the sample to jump into different modes. When the modes are isolated, the momentum variable cannot jump out of the current mode, for the interval between two modes has tremendous potential energy. Generally, objects tend to stay in the low energy places which refer to low probability regions. Although we can enlarge the momentum variable to pass through high potential energy places, the momentum variable should be exponential order large, which causes a decrease in the performance of the samplers rapidly. To deal with the problem of multi-modal sampling, several studies have been developed \citep{sminchisescu2007generalized,lan2014wormhole,Tripuraneni2017}. \citet{sminchisescu2007generalized} proposed a new dynamics sampler which is based on a darting algorithm \citep{andricioaei2001smart}. However, when the dimensions are high, this algorithm may have low efficiency. \citet{lan2014wormhole} uses the natural gradient of the target distribution to establish paths between different modes, and thus the samples can jump through the low probability regions. This method may suffer from low effective sample size (ESS) \citep{Neal2010}, which means that the relationship between two neighbor samples is close, so the samples are not independent. \citet{Tripuraneni2017} introduced the concept of magnetic filed. By means of constructing a dynamics system based on magnetic filed, this method can achieve great performance in multi-modal sampling. However, the setting of the magnetic field parameter is difficult and this method may also suffer from high autocorrelation in multi-modal sampling.

In this paper, we introduce a novel dynamics MCMC method which is called variational Langevin Hamiltonian Monte Carlo (VHMC). This method exploits the variational distribution \citep{blei2017variational} of the target distribution to help dynamics sampler to find the new mode. A new Metropolis-Hasting criterion is proposed to satisfy the detailed balance condition \citep{martino2013flexibility}. Since the variational distribution has the modes information, samples can exploit this information to jump between different modes. Because dynamics based samplers can sample unimodal distributions well and variational distribution guides the dynamics based sampler to jump between modes, VHMC can overcome the distant multi-modal sampling problem. Furthermore, we improve Hamiltonian dynamics through Langevin dynamics and equipotential conversion to reduce the autocorrelation of samples and accelerate the convergence of the dynamics sampler. Finally, a detailed proof is given to demonstrate that our algorithm can converge to the target distribution.

Both synthetic data and real data experiments are conducted to verify our theory. We sample points from 7 different Gaussian mixture distributions whose dimensions range from 2 to 256. We apply our method to two-class classification exploiting Bayesian logistic regression \citep{mackay1992evidence} to test the performance of VHMC on real datasets. Evaluation indices like maximum mean discrepancy \citep{gretton2012kernel} and autocorrelation are calculated to assess the quality of samples. Experiment results illustrate that the proposed method is capable of sampling from distant multi-modal distribution while obtaining better performance compared with other state-of-the-art methods \citep{zhang2016towards,Tripuraneni2017}.

The main contributions of this work can be summarized as follows. We propose a novel sampler called Langevin Hamiltonian Monte Carlo (LHMC), which achieves lower autocorrelation and faster convergence compared with HMC sampler. Since HMC sampler has poor performance in multi-modal sampling, we propose a new method, which utilizes the variational distribution of the target distribution to guide the sampler to jump through different modes. A detailed proof is given to prove the correctness of our method. Sufficient experiments on various multi-modal distributions with different dimensions and Bayesian logistic regression are conducted. We observe that the proposed method achieves better performance compared with other algorithms.

The rest of this article is organized as follows. In Section 2, we review the background of our study, including the introduction of Hamiltonian Monte Carlo and Langevin dynamics. In Section 3, we introduce our LHMC sampler and show the objective function of the proposed method. In Section 4, we propose the variational Hamilton Monte Carlo, which aims to address the problem of multi-modal sampling. In Section 5, a detailed proof to demonstrate the correctness of the algorithm is given. Experiments and analysis are given in Section 6. In Section 7, we conclude this paper and discuss future work.

\section{Background Knowledge}
In this section, we introduce the basic methods which we exploit in our method. First, we introduce Hamiltonian Monte Carlo, a widely used MCMC sampler in Bayesian machine learning. Our method is based on the HMC sampler. And then, Langevin dynamics is introduced. We use Langevin dynamics to improve the performance of HMC sampler.
\subsection{Hamiltonian Monte Carlo}
Hamiltonian Monte Carlo (HMC) \citep{duane1987hybrid,Neal2010} is one of the state-of-the-art Markov chain Monte Carlo algorithms. The use of gradient information of the target distribution makes HMC more efficient than the traditional Metropolis-Hasting (MH) algorithms . HMC has a high probability of acceptance, while MH algorithms employ the random walk scheme to explore the state space.

HMC exploits Hamiltonian dynamics to calculate the new state, whose state space is composed of joint Gaussian momentum $p$ and position $\theta$, where $p$ is independent of the variable $\theta$. Suppose we use a spring oscillator to describe the Hamiltonian dynamics. Then we can get the following equation:
\begin{equation}
H(\theta,p)=U(\theta)+K(p),
\end{equation}
where $U(\theta)$ represents the potential energy of the ball at the position $\theta$ at time $t$ while $K(p)$ represents the kinetic energy of the ball at time $t$. $H(p,\theta)$ represents the total energy of the ball. In order to construct Hamiltonian dynamics, the derivatives of position $\theta$ and momentum $p$ about time are required. The Hamiltonian equations are formed as follows:
\begin{equation}
\begin{aligned}
\frac{{\rm d}\theta}{{\rm d} t}&=\frac{\partial H(p,\theta)}{\partial p}=\nabla_p K(p)=M^{-1}p\\
\frac{{\rm d} p}{{\rm d} t}&=-\frac{\partial H(p,\theta)}{\partial \theta}=-\nabla_\theta U(\theta).
\end{aligned}
\label{HamEq2}
\end{equation}

In practice, it is difficult to get the exact solutions to these differential equations. HMC instead discretizes these equations through using non-zero time steps, which inevitably introduces some error. It is, nevertheless, necessary to use a discretization for which Liouville's theorem holds exactly \citep{neal1993probabilistic}. The common discretization method of HMC is leapfrog which takes the form as:
\begin{equation}
\begin{aligned}
p\left(t+\frac{\epsilon}{2}\right)&=p\left(t\right)-\frac{\epsilon}{2}\nabla_\theta U(\theta(t))\\
\theta\left(t+\epsilon\right)&=\theta\left(t\right)+\epsilon \nabla_p K\left(p\left(t+\frac{\epsilon}{2}\right)\right)\\
p\left(t+\epsilon\right)&=p\left(t+\frac{\epsilon}{2}\right)-\frac{\epsilon}{2}\nabla_\theta(\theta\left(t+\epsilon\right)),
\end{aligned}
\label{Hamleapfrog}
\end{equation}
where $\epsilon$ represents the step size. Leapfrog perserves the phase space volume and is also time reversible. Through the discretization method we can get the new state, and HMC methods then apply Metropolis-Hasting to the new state to decide whether accept or reject the state, which takes the form as:

\begin{equation}
\begin{aligned}
{\rm {min}}\left(1,\mathrm{exp}\left(H(\theta_{(n-1)},p_{(n-1)})-H(\hat{\theta},\hat{p})\right)\right),
\end{aligned}
\label{MH}
\end{equation}
where $\theta_{(n-1)},p_{(n-1)}$ represents the last state and $\hat{\theta},\hat{p}$ represents the newly proposed state. By means of controlling the leapfrog size $L$ and small step $\epsilon$, we can adjust the acceptance rate of the HMC sampler. Algorithm~\ref{alg:hmc} gives the pseudo code of HMC \citep{Neal2010}.

Suppose we need to sample from the distribution of $\theta$ given the observation data $\mathcal{D}$:
\begin{equation}
p(\theta|\mathcal{D})\propto e^{-U(\theta)},
\end{equation}
\begin{algorithm}[tb]
	\caption{Hamiltonian Monte Carlo}
	\label{alg:hmc}
	\begin{algorithmic}
		\STATE {\bfseries Input:} step size $\epsilon$, leapfrog size $L$, starting point $\theta^{(1)}$, sample number $N$
        \STATE {\bfseries Output:} Samples $\theta_{(1:N)}$
		\FOR{$n=1$  {\bfseries to} $N$}
		\STATE Resample the momentum variable $p$
		\STATE $p_{(n)}\sim \mathcal{N}(0,1)$
		\STATE $(\theta_{0},p_{0})=(\theta_{(n)},p_{(n)})$
		\STATE $p_{0}=p_{0}-\frac{\epsilon}{2}\nabla_\theta U(\theta_{0})$
		\STATE $\theta_{0}=\theta_{0}+\epsilon\nabla_p K(p_{0})$
		\FOR{$i=1$ {\bfseries to} $L$}
		\STATE $p_i=p_{i-1}-\epsilon \nabla_\theta U(\theta_{i-1})$
		\STATE $\theta_i=\theta_{i-1}+\epsilon\nabla_p K(p_i)$
		\ENDFOR
		\STATE $p_L=p_L-\frac{\epsilon}{2}\nabla_\theta U(\theta_L)$
		\STATE $(\hat{\theta},\hat{p})=(\theta_L,p_L)$
		\STATE Metropolis-Hasting procedure
        \STATE $u\sim {\rm Uniform}(0,1)$
		\STATE $\alpha = {\rm {min}}\left(1,\mathrm{exp}\left((U(\theta_{(n)})+K(p_{(n)}))-(U(\hat{\theta})+K(\hat{p}))\right)\right)$
		\IF{$\alpha > u$}
		\STATE $(\theta_{(n+1)}, p_{(n+1)})=(\hat{\theta}, \hat{p})$
		\ELSE
		\STATE $(\theta_{(n+1)}, p_{(n+1)})=(\theta_{(n)}, p_{(n)})$
		\ENDIF
		\ENDFOR
	\end{algorithmic}
\end{algorithm}
where we have the form of potential energy:
\begin{equation}
U(\theta)\propto -\rm ln p(\theta|\mathcal{D}).
\end{equation}
According to the Hamiltonian dynamics, through introducing a set of auxiliary momentum variables $p$, HMC sampler is able to sample the jointly distribution $\pi(\theta,p)$  defined as:
\begin{equation}
\pi(\theta,p)\propto e^{-H(\theta,p)} .
\end{equation}
Through using the Hamiltonian equations (\ref{HamEq2}), we get the new state of $\theta$ and $p$. Because the position variable $\theta$ and the momentum variable $p$ are independent, sampling $\theta$ and $p$ alternatively will not affect the results.

The Hamiltonian dynamics has three properties. First, it preserves the total energy $H(\theta_{(t)},p_{(t)})=H(\theta_{(0)},p_{(0)})$, and thus the joint probability density has $p(\theta_{(t)},p_{(t)})=p(\theta_{(0)},p_{(0)})$. Second, it preserves the volume element. Last, it is time reversible \citep{leimkuhler2004simulating}. As a result, if the potential energy and the kinetic energy remain unchanged during the dynamics system, then the joint probability density of $\theta$ and $p$ also remains unchanged.

Compared with the random walk strategy, HMC methods explore the target distribution much more efficiently due to the use of gradient information. HMC can travel a long distance in the phase space $(\theta,p)$, which enhance the acceptance rate. However, it is really difficult for HMC to travel across the low probability region in which the value of the gradient of the potential energy is very large. Enlarging the momentum variable $p$ may be helpful to jump over these regions, but the efficiency of HMC may decrease significantly.

Recently, some new developments of HMC have been proposed to make HMC sampler more flexible. For example, Riemann manifold HMC \citep{Girolami2011} exploits the Riemann geometry to tune the mass $M$, which tend to create a more efficient HMC sampler. The "No U-Turn" sampler \citep{hoffman2014no} can tune the step size $\epsilon$, leapfrog length $L$ and the simulation steps automatically. We note that in principle these state-of-the-art HMC samplers can also be combined with our proposed method.

\subsection{Langevin dynamics}
Langevin dynamics was first utilized to describe the diffusion process of molecular systems. MCMC samplers based on Langevin dynamics \citep{brunger1984stochastic,burrage2009accurate,milstein2013stochastic} have already been proposed. Langevin dynamics is a system of Ito-type stochastic differential equations, which takes the form as:
\begin{equation}
\begin{aligned}
\mathrm{d}\theta=-M^{-1}\nabla U\left(\theta\right)\mathrm{d}t+\sqrt{\frac{2\beta^{-1}}{M}}\mathrm{d}W,
\label{langevin_dynamic0}
\end{aligned}
\end{equation}
where $W$ represents the stochastic Wiener process, $M$ represents the diagonal mass matrix, $U(\theta)$ represents the energy function and $\beta^{-1}=k_{B}T$, where $k_{B}$ is Boltzmann’s constant and $T$ represents the temperature. Since solving (\ref{langevin_dynamic0}) is difficult, Euler-Maruyama is used to approximately solve the differential equation, which takes the form as:
\begin{equation}
\begin{aligned}
\theta_{n+1}=\theta_{n}-\frac{\epsilon^{2}}{2}M^{-1}\nabla U\left(\theta_{n}\right)+\sigma \epsilon^{2}\sqrt{\frac{2\beta^{-1}}{M}}z_{n},
\label{discretization_langevin_dynamic0}
\end{aligned}
\end{equation}
where $z\sim \mathcal{N}(z|0,\textbf{I})$, $\epsilon$ represents the integration step size. (\ref{discretization_langevin_dynamic0}) only gives the solution to the overdamped Langevin dynamics \citep{leimkuhler2012rational}, which means that the friction term has not been concerned. Next we talk about the Langevin dynamics with the friction term \citep{leimkuhler2013robust}, which is defined as:
\begin{equation}
\begin{aligned}
\mathrm{d}\theta&=M^{-1}p\mathrm{d}t\\
\mathrm{d}p&=-\nabla U\left(\theta\right)\mathrm{d}t-\gamma p \mathrm{d}t+\sigma \sqrt{M}\mathrm{d}W,
\label{langevin_dynamic}
\end{aligned}
\end{equation}
where $\gamma>0$ represents the friction factor and $\sigma=\sqrt{2\gamma\beta^{-1}}$. However, simulating (\ref{langevin_dynamic}) is very difficult, so discretization method \citep{leimkuhler2012rational} is utilized to solve the above stochastic differential equation, which takes the form as:
\begin{equation}
\begin{aligned}
p_{(n+1)/2}&=p_n-\frac{\epsilon}{2}\nabla U\left(\theta_n\right)\\
\theta_{(n+1)/2}&=\theta_n+\frac{\epsilon}{2}M^{-1}p_{(n+1)/2}\\
\hat{p}_{(n+1)/2}&=a_{1}p_{(n+1)/2}+a_2\sqrt{M}z_{n+1}\\
\theta_{n+1}&=\theta_{(n+1)/2}+\frac{\epsilon}{2} M^{-1} \hat{p}_{(n+1)/2}\\
p_{n+1}&=\hat{p_{(n+1)/2}}-\frac{\epsilon}{2}\nabla U\left(\theta_{n+1}\right),
\label{discretization_langevin_dynamic}
\end{aligned}
\end{equation}
where $z\sim \mathcal{N}(z|0,\textbf{I})$, $\epsilon$ represents the step size and $a_{1}=\mathrm{e}^{-\gamma\epsilon}$ and $a_2=\sqrt{\beta^{-1}\left(1-a_{1}^{2}\right)}$.

Compared with Hamiltonian dynamics, Langevin dynamics allows exploring the state space more freely, for Langevin dynamics concern about the friction between the molecule and the thermal motion of the molecule, which constructs a more real environment.

\section{Langevin Hamiltonian Monte Carlo}

HMC exploits the Hamiltonian dynamics to propose the new sample. However, HMC sampler may have large autocorrelation because each new sample is obtained through the deterministic calculation of the last sample. Specifically, (\ref{Hamleapfrog}) defines the process of calculating new state $\left(\theta_{t+1},p_{t+1}\right)$ through the old state $\left(\theta_{t},p_{t}\right)$. It is the deterministic computation by using the gradient information that causes the high autocorrelation of the HMC sampler.

In order to reduce the autocorrelation of the HMC sampler, we propose the Langevin Hamiltonian Monte Carlo (LHMC). The main idea of LHMC is to take advantage of Langevin dynamics to add the randomness to the proposed state and introduce the concept of equipotential transformation (ET) for some individual cases. The difference between Hamiltonian dynamics and Langevin dynamics is that Langevin dynamics provides a random walk of momentum variable while the total energy remains unchanged. In Langevin dynamics, we consider that the total energy consists of the potential energy, kinetic energy and internal energy, which takes the form as:
\begin{equation}
\begin{aligned}
H=U(\theta)+K(p)+Q,
\label{toatal_energy}
\end{aligned}
\end{equation}
where $Q$ represents the internal energy. The random thermal motion consumes the internal energy which finally transforms into the kinetic energy, which is described as (\ref{internal_energy}):
\begin{equation}
\begin{aligned}
p_t&=a_{1}p_{t-1}+a_2\sqrt{M}z\\
\Delta E&=K(p_t)-K(p_{t-1})\\
Q_t&=Q_{t-1}-\Delta E.
\label{internal_energy}
\end{aligned}
\end{equation}
We use Metropolis-Hasting criterion to accept the samples, so the acceptance rate $\varphi$ takes the form as:
\begin{equation}
\begin{aligned}
\varphi&=min\left(1,\mathrm{exp}\left((U_{t-1}+K_{t-1}+Q_{t-1})-(U_{t}+K_{t}+Q_{t})\right)\right)\\
&=min\left(1,\mathrm{exp}\left((U_{t-1}+K_{t-1}+Q_{t-1})-(U_{t}+K_{t}+Q_{t-1}-\Delta E)\right)\right)\\
&=min\left(1,\mathrm{exp}\left((U_{t-1}+K_{t-1})-(U_{t}+K_{t}-\Delta E)\right)\right).
\label{lhmh}
\end{aligned}
\end{equation}

Inspired by (\ref{langevin_dynamic}), we propose the equipotential transformation of the potential energy. Suppose the probability density function is symmetrical which is denoted as $p(\theta)$, then the potential energy function can be written as: $U(\theta) = -{\rm{ln}}p(\theta)$. If we want to obtain the equipotential state of $\theta_0$, we should calculate $U(\hat{\theta})=U(\theta_0)$ and that is to solve $U(\hat{\theta})-U(\theta_0)=0$. Let $f(\theta,\theta_{(n)})=U(\hat{\theta})-U(\theta_{(n)})$ , $f^{'}(\theta,\theta_{(n)})=\nabla_\theta U(\theta)-\nabla_\theta U(\theta_{(n)})$ and $\hat{\theta_0}=\mathcal{N}(\theta|\theta_{(n)},\sigma)$. We iteratively solve the equipotential state through (\ref{newton}).
\begin{equation}
\begin{aligned}
\hat{\theta_t} = \hat{\theta_{t-1}}-\frac{f(\hat{\theta_{t-1}},\theta_{(n)})}{f^{'}(\hat{\theta_{t-1}},\theta_{(n)})}, t=1...T,
\label{newton}
\end{aligned}
\end{equation}
where $T$ represents the iteration times.

\begin{figure} \includegraphics[width=0.4\columnwidth,height=6cm]{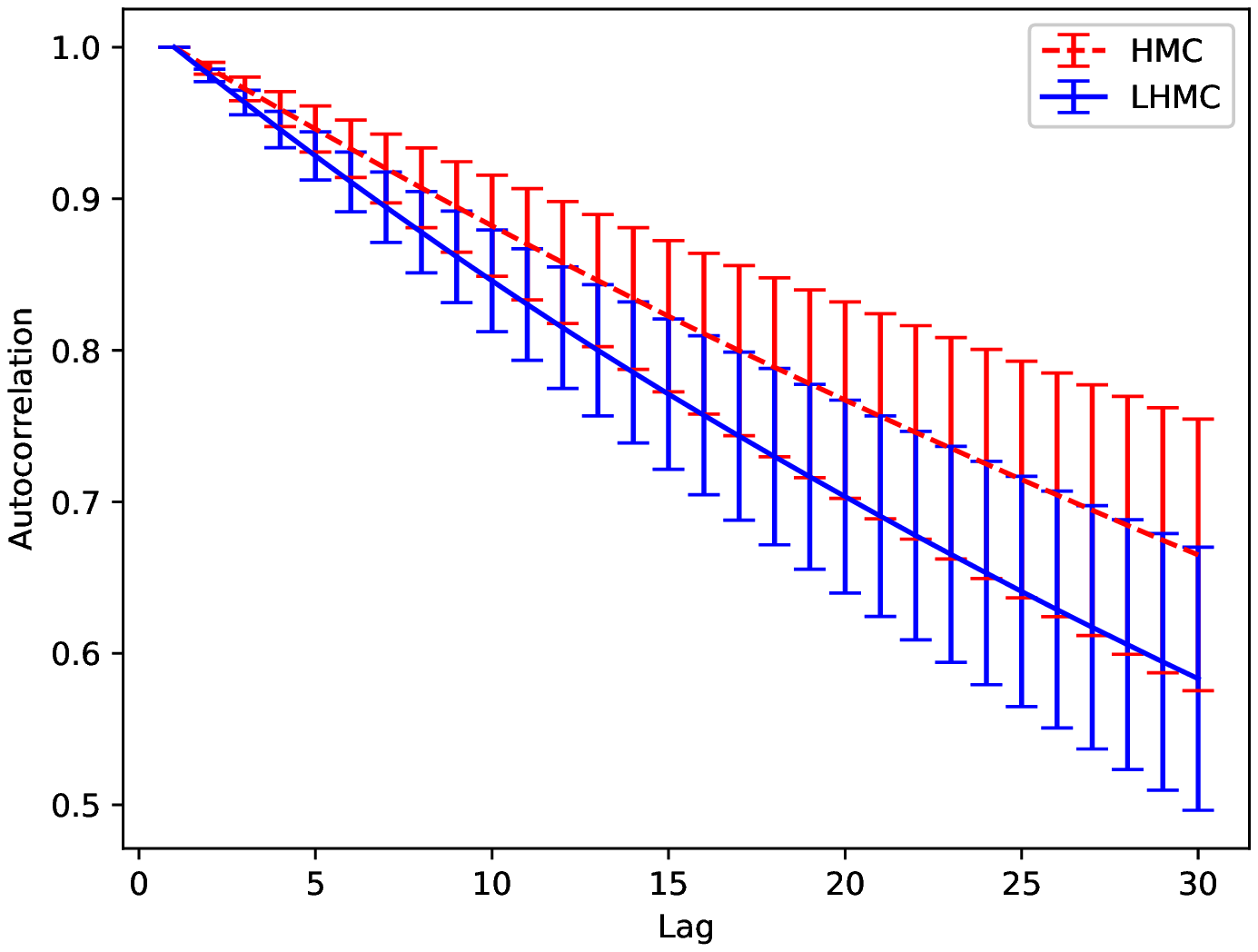} \includegraphics[width=0.4\columnwidth,height=6cm]{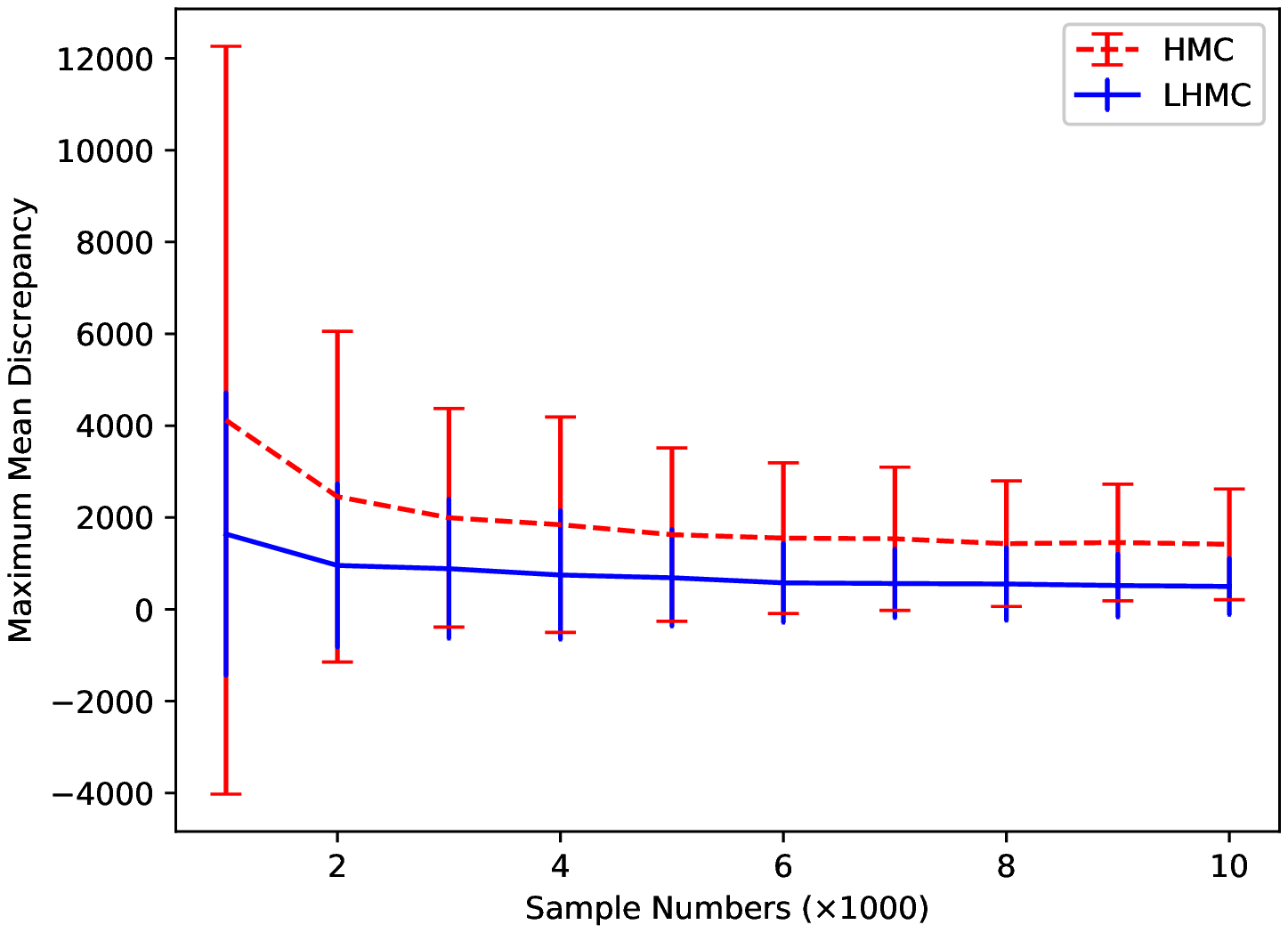}
\includegraphics[width=0.4\columnwidth,height=6cm]{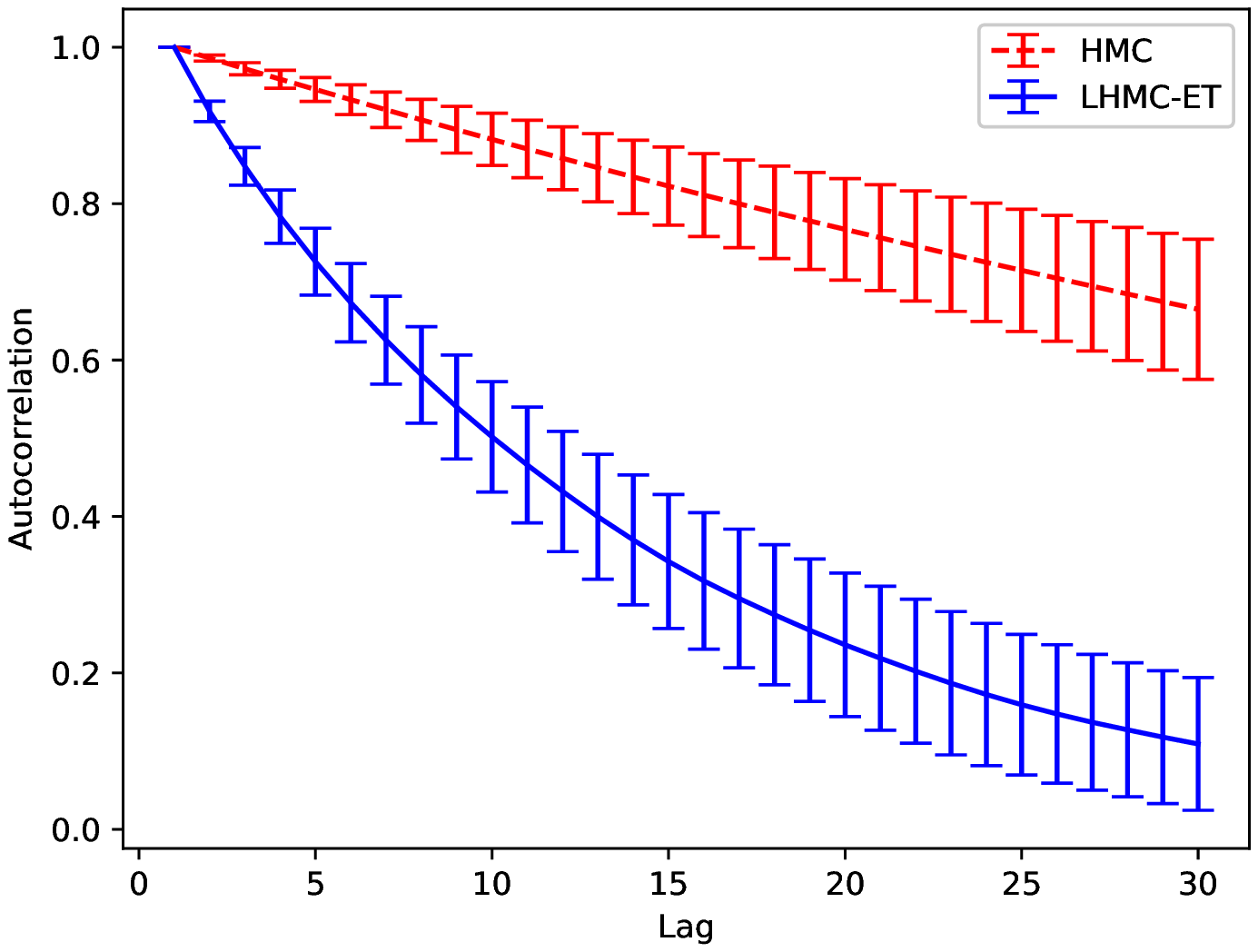} \includegraphics[width=0.4\columnwidth,height=6cm]{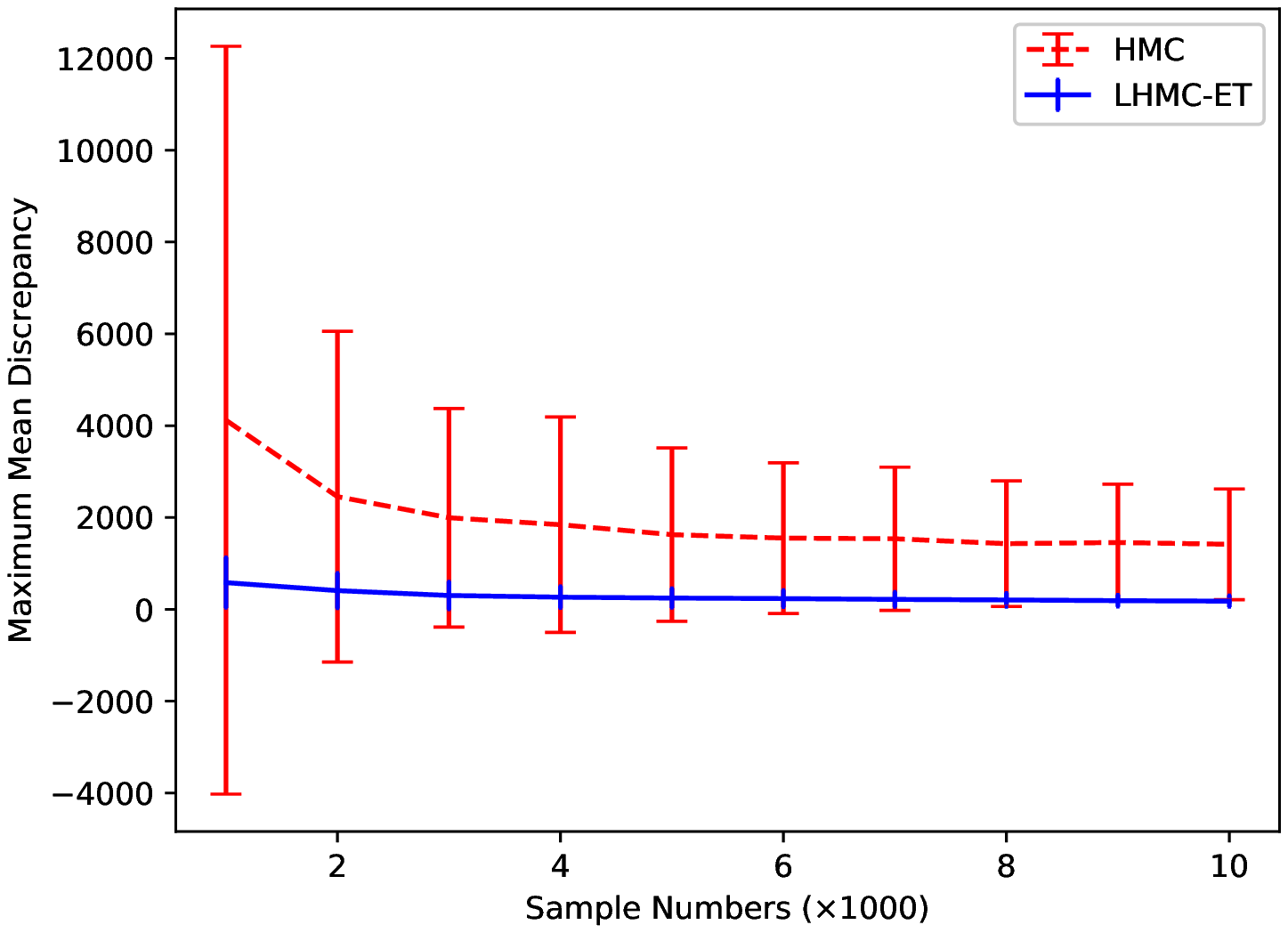}
\centering
\caption{The comparison of MMD and autocorrelation for three different methods. The left column shows the relationship between autocorrelation and lag of sample number. The right column demonstrates the relationship between MMD and sample number (best viewed in color).}
\label{mmd-auto}
\end{figure}

Given the target distribution, LHMC exploits Langevin dynamics and Hamiltonian dynamics to propose the new sample. In addition, LHMC may provide the equipotential transformation in the process of proposing the new state for some symmetrical distributions. LHMC can be summarized as three stages. The first stage is Langevin dynamics, which takes the form as (\ref{langevin_dynamic}). The second stage is Hamiltonian dynamics, which takes the form as (\ref{HamEq2}) and the last stage is also Langevin dynamics. Assume the initial state is $(\theta,p)$, a half update of the Langevin dynamics can be written as:
\begin{equation}
\begin{aligned}
p_{(n+1)/2}&=p_n-\frac{\epsilon}{2}\nabla U\left(\theta_n\right)\\
\theta_{(n+1)/2}&=\theta_n+\frac{\epsilon}{2}M^{-1}p_{(n+1)/2}.
\label{1half_langevin_dynamic}
\end{aligned}
\end{equation}
The random thermal motion of molecules takes the form as:
\begin{equation}
\begin{aligned}
\hat{p}_{(n+1)/2}&=a_{1}p_{(n+1)/2}+a_2\sqrt{M}z_{n+1}.
\label{thermal}
\end{aligned}
\end{equation}
The other half update of the Langevin dynamics can be written as:
\begin{equation}
\begin{aligned}
\theta_{n+1}&=\theta_{(n+1)/2}+\frac{\epsilon}{2} M^{-1} \\
p_{n+1}&=\hat{p_{(n+1)/2}}-\frac{\epsilon}{2}\nabla U\left(\theta_{n+1}\right),
\label{2half_langevin_dynamic}
\end{aligned}
\end{equation}
The detailed algorithms of LHMC-ET and LHMC are described in Algorithm~\ref{alg:lhmc_et} and Algorithm~\ref{alg:lhmc} respectively. We demonstrate the performance of LHMC and LHMC-ET on a strongly correlated Gaussian, which has the symmetrical PDF. A diagonal Gaussian with variances $[10^2,10^{-2}]$ is rotated by $\frac{\pi}{4}$, which is an extreme circumstance of \citet{Neal2010}. We experiment 100 times and calculate the mean and variance of autocorrelation and maximum mean discrepancy. We set $M=1.2\textbf{I}$, $\gamma=0.5$, $T=200$, $\epsilon=0.05$, leapfrog size $L=40$, equipotential transformation iteration length $EL=10$, and $err=0.01$. As Figure~\ref{mmd-auto} illustrates, LHMC achieves lower autocorrelation and faster convergence rate compared with HMC. Besides, for this symmetrical distribution, LHMC-ET achieves lower autocorrelation than LHMC, which indicates that the random walk of momentum variable and ET do help us to design a powerful and better sampler.

\begin{algorithm}[tb]
	\caption{Langevin Hamiltonian Monte Carlo with ET (LHMC-ET)}
	\label{alg:lhmc_et}
	\begin{algorithmic}
		\STATE {\bfseries Input:} step size $\epsilon$, leapfrog length $L$, equipotential transformation iteration length $EL$, error of equipotential transformation $err$, starting point $\theta_{(1)}$, sample number $N$,
        \STATE {\bfseries Output:} Samples $\theta_{(1:N)}$
		\FOR{$n=1$  {\bfseries to} $N$}
        \STATE $u\sim {\rm Uniform}(0,1)$
        \STATE $U_{error}=0$
        \IF{$\alpha > u$}
        \STATE $x\sim \mathcal{N}(\theta_{n},\sigma)$
        \FOR{$i=1$  {\bfseries to} $EL$}
        \STATE $x=x-\frac{f(x,\theta_{n})}{f^{'}(x,\theta_{n})}$
        \ENDFOR
        \IF{$-err<U_{error}<err$}
        \STATE $\theta^{'} =x$
        \ENDIF
        \STATE Resample the momentum variable $p$
        \STATE $\hat{\theta},\hat{p},\Delta E=DLHMC(\theta^{'},p,\epsilon,L)$
        \ELSE
        \STATE Resample the momentum variable $p$
        \STATE $\hat{\theta},\hat{p},\Delta E=DLHMC(\theta^{n},p,\epsilon,L)$
        \STATE $x\sim \mathcal{N}(\hat{\theta},\sigma)$
        \FOR{$i=1$  {\bfseries to} $EL$}
        \STATE $x=x-\frac{f(x,\hat{\theta})}{f^{'}(x,\hat{\theta})}$
        \ENDFOR
        \IF{$-err<U_{error}<err$}
        \STATE $\hat{\theta} =x$
        \ENDIF
        \ENDIF
		\STATE Metropolis-Hasting procedure
        \STATE $u\sim {\rm Uniform}(0,1)$
		\STATE $\alpha = {\rm {min}}\left(1,\mathrm{exp}\left(U(\theta_{(n-1)}+K(p_{(n-1)}))-(U(\hat{\theta})+K(\hat{p})-\Delta E)\right)\right)$
		\IF{$\alpha > u$}
		\STATE $(\theta_{(n+1)}, p_{(n+1)})=(\hat{\theta}, \hat{p})$
		\ELSE
		\STATE $(\theta_{(n+1)}, p_{(n+1)})=(\theta_{(n)}, p_{(n)})$
		\ENDIF
		\ENDFOR
	\end{algorithmic}
\end{algorithm}

\begin{algorithm}[tb]
	\caption{Langevin Hamiltonian Monte Carlo (LHMC)}
	\label{alg:lhmc}
	\begin{algorithmic}
		\STATE {\bfseries Input:} step size $\epsilon$, leapfrog size $L$, starting point $\theta_{(1)}$, sample number $N$
        \STATE {\bfseries Output:} Samples $\theta_{(1:N)}$
		\FOR{$n=1$  {\bfseries to} $N$}
        \STATE Resample the momentum variable $p$
        \STATE $\hat{\theta},\hat{p},\Delta E=DLHMC(\theta_{n},p,\epsilon,L)$

		\STATE Metropolis-Hasting procedure
        \STATE $u\sim {\rm Uniform}(0,1)$
		\STATE $\alpha = {\rm {min}}\left(1,\mathrm{exp}\left(U(\theta_{(n-1)}+K(p_{(n-1)}))-(U(\hat{\theta})+K(\hat{p})-\Delta E)\right)\right)$
		\IF{$\alpha > u$}
		\STATE $(\theta_{(n+1)}, p_{(n+1)})=(\hat{\theta}, \hat{p})$
		\ELSE
		\STATE $(\theta_{(n+1)}, p_{(n+1)})=(\theta_{(n)}, p_{(n)})$
		\ENDIF
		\ENDFOR
	\end{algorithmic}
\end{algorithm}

\begin{algorithm}[tb]
	\caption{Discretization for Langevin Hamiltonian Monte Carlo (DLHMC)}
	\label{alg:d_lhmc}
	\begin{algorithmic}
		\STATE {\bfseries Input:} step size $\epsilon$, leapfrog size $L$, starting point $(\theta_{(n)},p_{(n)})$
        \STATE {\bfseries Output:} $\theta_{(n+1)}$, $\theta_{(n+1)}$, $\Delta E$
        \STATE $\Delta E=0$
        \STATE Obtaining $\theta_{(n+1)/6},p_{(n+1)/6}$ through (\ref{1half_langevin_dynamic}).
        \STATE $\hat{p}=p_{(n+1)/6}$
        \STATE Obtaining the thermal motion of molecules $p_{(n+1)/6}$ through (\ref{thermal}).
		\STATE $\Delta E= \Delta E+K(p_{(n+1)/6})-K(\hat{p})$
        \STATE Obtaining $\theta_{(n+1)/3},p_{(n+1)/3}$ through (\ref{2half_langevin_dynamic}).
        \STATE Obtaining $\theta_{2(n+1)/3},p_{2(n+1)/3}$ by simulating Hamiltonian dynamics through (\ref{Hamleapfrog}).
        \STATE Obtaining $\theta_{5(n+1)/6},p_{5(n+1)/6}$ through (\ref{1half_langevin_dynamic}).
        \STATE $\hat{p}=p_{5(n+1)/6}$
        \STATE Obtaining the thermal motion of molecules $p_{5(n+1)/6}$ through (\ref{thermal}).
		\STATE $\Delta E= \Delta E+K(p_{5(n+1)/6)}-K(\hat{p})$
        \STATE Obtaining $\theta_{(n+1)},p_{(n+1)}$ through (\ref{2half_langevin_dynamic}).
	\end{algorithmic}
\end{algorithm}

\section{Variational Langevin Hamiltonian Monte Carlo}
It is acknowledged that the HMC methods can not well sample from multi-modal distributions when the modes are far away from each other especially in high dimensions \citep{lan2014wormhole, Tripuraneni2017}. In this section, we present a novel MCMC method called variational Hamiltonian Monte Carlo (VHMC) and prove that it is able to target the correct distribution.

\subsection{Variational Langevin Hamiltonian Monte Carlo}
Recent studies \citep{lan2014wormhole, Tripuraneni2017} have shown that MCMC samplers based on dynamics are challenging to deal with multi-modal distributions since there exist low probability regions between the isolated modes. Once the initial point is chosen, these methods could only sample from one mode. Nevertheless, if we only consider one mode, the samplers based on dynamics can sample the target distribution well.

\begin{figure}[htbp]
\centering
\includegraphics[width=12cm,height=6cm]{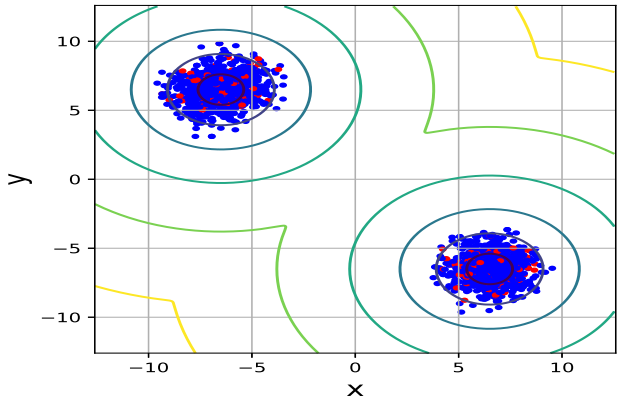}
  \caption{The guide points (red points) when using VHMC (best viewed in color).}
  \label{guide}
\end{figure}

In this study, we are aiming to address the problem of multi-modal sampling. A novel MCMC algorithm is proposed. We propose a new concept referred to guide points which are illustrated in Figure~\ref{guide}. Guide points are samples generated from the variational distribution. With the help of these guide points, VHMC is able to travel across the low probability regions between two modes.

Previous work \citep{lan2014wormhole, Tripuraneni2017} has already demonstrated that HMC sampler is difficult to sample the multi-modal distributions. However, even when the dimensions are high, HMC is capable of sampling single mode distributions, and thus we take advantage of these gradient based samplers to establish our sampler. Suppose we want to sample from the distribution $p(\theta)$. In order to get the local optimum solutions, $N$ samples are generated from the solution space, and we then use these $N$ initial points with Adam \citep{kingma2014adam} to calculate the optimum solution. From these optimum solutions, we can get $k$ modes. For each mode, we use the optimum solution as the initial state of LHMC sampler to generate $M$ samples. After that we can obtain the samples set $S_i=\{\theta|\theta\sim p_i(\theta)\},i=1...k$, which are generated in the single mode distribution $p_i(\theta),i=1...k$. We assume that each mode follows Gaussian distribution $q_i(\theta),i=1...k$. As a result, our purpose is to utilize $q_i(\theta)$ to approximate $p_i(\theta)$, where we use the $\mathbb{KL}$ divergence to quantify the similarity of the two distributions:
\begin{equation}
\begin{aligned}
\mathbb{KL}(p_i(\theta)||q_i(\theta))&=\int{p_i(\theta){\mathrm{ln}}\frac{p_i(\theta)}{q_i(\theta;\mu,\Sigma)}\rm{d}\theta}=E_{p_i(\theta)}[{\mathrm{ln}}\frac{p_i(\theta)}{q_i(\theta;\mu,\Sigma)}].
\label{KL}
\end{aligned}
\end{equation}
Since the integrator in (\ref{KL}) is difficult to calculate, Monte Carlo approximation is used to calculate the $\mathbb{KL}$ divergence. The integrator can be calculated as: $\frac{1}{M}\sum_{n=1}^{M}{{\mathrm{ln}}\frac{p_i(\theta_n)}{q_i(\theta_n;\mu,\Sigma)}}$, where $\theta_n \sim p_i(\theta)$.
Since $S_i$ is fixed and $p_i(\theta_n)$ is difficult to calculate, minimizing $\mathbb{KL}(p_i(\theta)||q_i(\theta))$ can be simplified as:

\begin{equation}
\begin{aligned}
{\rm{min}}\mathbb{KL}{(p_i(\theta)||q_i(\theta))}&={\rm{min}}~\biggl[\int {p_i(\theta){\mathrm{ln}}p_i(\theta)\rm{d}\theta}-\int {p_i(\theta){\mathrm{ln}}q_i(\theta;\mu,\Sigma)\rm{d}\theta}]\biggr]\\
&={\rm{min}}~\biggl[-\int {p_i(\theta){\mathrm{ln}}q_i(\theta;\mu,\Sigma)\rm{d}\theta}\biggr]\\
&={\rm{max}}~\biggl[E_{p_i(\theta)}[{\mathrm{ln}}q_i(\theta;\mu,\Sigma)]\biggr].
\label{min_kl}
\end{aligned}
\end{equation}

Using (\ref{min_kl}), we find that minimizing the $\mathbb{KL}$ divergence is equivalent to maximizing the likelihood. After getting parameters for each mode through maximizing the likelihood, we can obtain a mixture of Gaussian distribution as the variational distribution of the actual distribution. We generate new samples with two strategies. A vast amount of samples are generated with probability $1-\beta$ through LHMC sampler, while a few samples with probability $\beta$ are generated through the variational distribution, which is accepted with probability $\frac{p(\theta)}{cq(\theta)}$, where $c$ is a constant.

In LHMC sampler, we found that in Metropolis-Hasting procedure, the sampler rejects the proposed sample with probability $1-\alpha$ where $\alpha$ is defined in (\ref{lhmh}). This is an interesting phenomenon. If the proposed sample is rejected, it means $\varphi$ is smaller than 1, where $\theta_{n-1}$ represents the sample which is sampled at last step and $\hat{\theta}$ represents the newly proposed sample. In other words, $U(\hat{\theta})+K(\hat{p})-\Delta E$ must be much more larger than $U(\theta_{(n-1)})+K(p_{(n-1)})$. Samples with high probability have low potential energy and higher kinetic energy. If the proposed sample is rejected, the given kinetic energy $p_{n-1}$ must be large, and it will convert into potential energy. To understand this situation, let us think about a ball rolling in a "U" type surface. If the kinetic energy is given appropriately, the ball will always roll in the "U" type surface. However, if the kinetic energy is tremendous, the ball will jump out of the "U" type surface. When facing this situation, MCMC sampler would put the ball back to the last position. In our study, according to the sample proposed in the variational distribution, we put the ball into the proposed position. That is to say, if one sample is rejected in MH step, we generate a new sample from $q(\theta)$ which is accepted with the probability $\frac{p(\theta)}{cq(\theta)}$. Though the newly proposed acceptance rate $min(1,\frac{1-r(\theta^{*})}{1-r(\theta)})$, the detailed balance holds, where $r(.)$ represents the rejection probability of the current state jumping into other states. The detailed description is given in the convergence analysis part. The detailed algorithm is given in Algorithm~\ref{alg:vhmc}.

\begin{algorithm}
    \caption{Variational Hamiltonian Monte Carlo (VHMC)}
    \label{alg:vhmc}
    \begin{algorithmic}
        \STATE {\bfseries Input:} step size $\epsilon$, leapfrog size $L$, starting point $\theta_{(1)}$, sample number $N$
        \STATE {\bfseries Output:} Samples $\theta_{(1:N)}$
        \FOR{$n=1$  {\bfseries to} $N$}
        \STATE $u\sim Uniform(0,1)$
        \IF{$\beta < u$}
        \STATE Resample the momentum variable $p$
        \STATE $\hat{\theta},\hat{p},\Delta E=DLHMC(\theta_{n},p,\epsilon,L)$
        \STATE Metropolis-Hasting procedure
        \STATE $u\sim Uniform(0,1)$
        \STATE $\alpha = \rm {min}\left(1,\mathrm{exp}\left(U(\theta_{(n)}+K(p_{(n)}))-(U(\hat{\theta})+K(\hat{p})-\Delta E)\right)\right)$
        \IF{$\alpha > u$}
        \STATE $\theta_{(n+1)}=\hat{\theta}$
        \ELSE
        \STATE Sample $\theta^{*}$ from the variational distribution $q$
        \STATE $h\sim Uniform(0,1)$
        \WHILE{$h>\frac{p(\theta^{*})}{xq(\theta^{*})}$}
        \STATE sample $\theta^{*}$ from the variational distribution q
        \ENDWHILE
        \STATE New Metropolis-Hasting procedure
        \STATE $u\sim {\rm{Uniform}}(0,1)$
        \IF{$u<\min(1,\frac{1-r(\theta^{*})}{1-\alpha})$}
        \STATE $\theta_{(n+1)}=\theta^{*}$
        \ELSE
        \STATE $\theta_{(n+1)}=\theta_{(n)}$
        \ENDIF
        \ENDIF
        \ELSE
        \STATE Sample $\theta^{*}$ from the variational distribution $q$
        \STATE $h\sim Uniform(0,1)$
        \WHILE{$h>\frac{p(\theta^{*})}{cq(\theta^{*})}$}
        \STATE sample $\theta^{*}$ from the variational distribution $q$
        \ENDWHILE
        \STATE $\theta^{(n+1)}=\theta^{*}$
        \ENDIF
        \ENDFOR
    \end{algorithmic}
\label{alg:vhmc}
\end{algorithm}
\begin{figure}
\centering
\includegraphics[width=10cm,height=8cm]{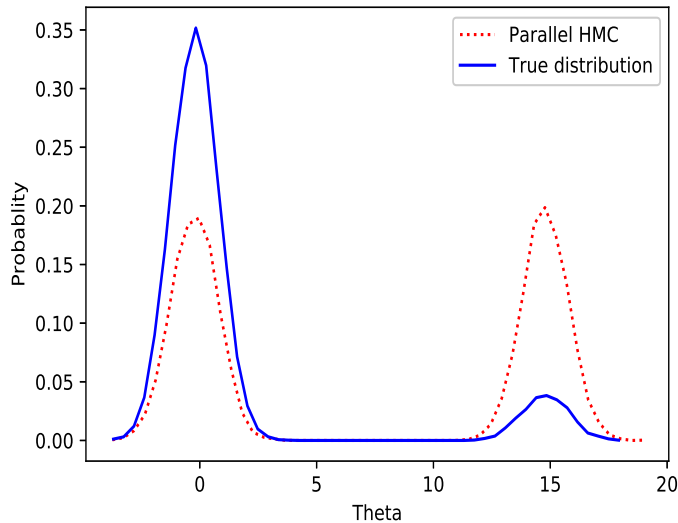}
\caption{The probability density function between parallel HMC and the actual distribution.}
\label{hmcpapallel}
\end{figure}
\subsection{Deficiency of Parallel HMC}
Although we can run $N$ HMC samplers in parallel to approximately sample from a multi-modal distribution, in high dimensions this kind of method is inaccurate. In other words, the probability of each mode may be the same, which can not reflect the actual distribution. Figure~\ref{hmcpapallel} shows that parallel HMC cannot sample from the actual distribution. The problem of parallel HMC is that the gradient direction cannot determine the probability of each mode.

\section{Convergence Analysis}
The correctness of VHMC will be proved in two aspects. First of all, we prove that $\pi(\theta,p)\propto\exp(-U(\theta))$ is the unique stationary distribution of the dynamics described in Algorithm~\ref{alg:lhmc_et}. For the symmetrical target distribution, the transformation probability of LHMC-ET can be written as:
\begin{equation}
\begin{aligned}
T(\theta_j,p_j|\theta_i,p_i)&=\frac{1}{2}p_{et}(\theta_{i}^{*}|\theta_i)T(\theta_2,p_2|\theta_{i}^{*},p_i)p(p_{2}^{*}|p_2)T(\theta_5,p_5|\theta_2,p_2^{*})p(p_5^{*}|p_5)T(\theta_j,p_j|\theta_5,p_{5}^{*}),\\
T(\theta_i,p_i|\theta_j,p_j)&=\frac{1}{2}T(\theta_5,p_{5}^{*}|\theta_j,p_j)p(p_5|p_5^{*})T(\theta_2,p_2^{*}|\theta_5,p_5)p(p_2|p_{2}^{*})T(\theta_{i}^{*},p_i|\theta_2,p_2)p_{et}(\theta_i|\theta_{i}^{*})\\
T(\theta_5,p_5|\theta_2,p_2^{*})&=T(\theta_3,p_3|\theta_2,p_2^{*})T(\theta_4,p_4|\theta_3,p_3)T(\theta_5,p_5|\theta_4,p_4),\\
T(\theta_2,p_2^{*}|\theta_5,p_5)&=T(\theta_4,p_4|\theta_5,p_5)T(\theta_3,p_3|\theta_4,p_4)T(\theta_2,p_2^{*}|\theta_3,p_3),
\label{rf}
\end{aligned}
\end{equation}
where $p_{et}(\theta_{i}^{*}|\theta_i)$ represents the transformation probability of equipotential conversion. Since the initial points are generated from the symmetry Gaussian distribution and the target distribution is symmetrical, we have $p_{et}(\theta_i|\theta_{i}^{*})=p_{et}(\theta_{i}^{*}|\theta_i)$. $T(\theta_2,p_2|\theta_{i}^{*},p_i)$ represents the transformation process defines in (\ref{1half_langevin_dynamic}), $p(p_{2}^{*}|p_2)$ represents the transformation process defines in (\ref{thermal}), $T(\theta_5,p_5|\theta_2,p_2^{*})$ represents the transformation process defines in (\ref{2half_langevin_dynamic}), (\ref{Hamleapfrog}) and (\ref{1half_langevin_dynamic}). $p(p_5^{*}|p_5)T(\theta_j,p_j|\theta_5,p_{5}^{*})$ represents the transformation process defines in (\ref{thermal}), (\ref{2half_langevin_dynamic}) respectively. Since the transformation processes define in (\ref{1half_langevin_dynamic}), (\ref{2half_langevin_dynamic}) and (\ref{Hamleapfrog}) are reversible, so $T(\theta_5,p_5|\theta_2,p_2^{*})=T(\theta_2,p_2^{*}|\theta_5,p_5)$. Finally $T(\theta_j,p_j|\theta_i,p_i)$ and $T(\theta_i,p_i|\theta_j,p_j)$ can be simplified as:
\begin{equation}
\begin{aligned}
T(\theta_j,p_j|\theta_i,p_i)&=c\cdot p(p_{2}^{*}|p_2)p(p_5^{*}|p_5)=c\cdot \mathcal{N}(a)\mathcal{N}(b),\\
T(\theta_i,p_i|\theta_j,p_j)&=c\cdot p(p_5|p_5^{*})p(p_2|p_{2}^{*})=c\cdot \mathcal{N}(b)\mathcal{N}(a),\\
\label{rf_simp}
\end{aligned}
\end{equation}
where $\mathcal{N}(.)$ is the Gaussian distribution. It is the symmetrical structure makes $T(\theta_j,p_j|\theta_i,p_i)=T(\theta_i,p_i|\theta_j,p_j)$.
Furthermore, since the total energy remains unchanged which means that $\frac{{\rm d}\pi(\theta,p)}{{\rm d}t}=0$, so we can verify that $\pi(\theta,r)$ is invariant, and we can imply that $\pi$ is a stationary distribution. However, $\pi$ is a stationary distribution under one model. It is not a global invariant distribution. Let us consider the detailed balance:

\begin{equation}
\pi(i)T(i,j)=\pi(j)T(j,i).
\label{DetailBalance}
\end{equation}
If (\ref{DetailBalance}) is satisfied, then $\pi$ is a stationary distribution. It is clear that in VHMC, $T(i,j)$ is not a symmetric distribution and it can be calculated as:
\begin{equation}
\begin{aligned}
T(i,j)&=r(i,i_r)\pi(j),\\
T(j,i)&=r(j,j_r)\pi(i),
\label{rf}
\end{aligned}
\end{equation}
where $r(i,i_r)$ represents rejection rate of $i$ to $i_r$, where $i_r$ is a rejected sample and $r(j,j_r)$ represents rejection rate of $j$ to $j_m$. In VHMC, we get this probability by calculating the probability rejected by Langevin Hamiltonian dynamics with the previous state. It is clear that HMC method will not jump into other modes when the modes are far away from each other, and the probability of jumping out of one mode can be calculated as $r(.)$. Nevertheless, the detailed balance in (\ref{DetailBalance}) is not satisfied. In VHMC, a further MH is introduced to keep the detailed balance which takes the form as:
\begin{equation}
\begin{aligned}
\pi(i)T(i,j)r(j,j_r)\pi(i)=\pi(j)T(j,i)r(i,i_r)\pi(j).
\label{fdb}
\end{aligned}
\end{equation}
As a result, the accept rate can be $r(j,j_r)\pi(i)$. In order to enlarge the accept rate, we enlarge the whole equation and make the accept rate to be $\phi={\rm{min}}\left[1,\frac{\pi(j)r(j,j_r)\pi(i)}{\pi(i)r(i,i_r)\pi(j)}\right]$. So the new detailed balance can be written as:

\begin{equation}
\begin{aligned}
\pi(i)T(i,j)\phi&=\pi(i)r(i,i_r)\pi(j)\phi\\
&={\rm{min}}\left[1,\frac{r(j,j_r)}{r(i,i_r)}\right]\pi(i)r(i,i_r)\pi(j)\\
&=\pi(j)r(j,j_r)\pi(i)\\
&=\pi(j)T(j,i).
\label{db}
\end{aligned}
\end{equation}
Finally, we prove that $\pi$ is an invariant distribution.

\section{Experiments}
In this section, we investigate the performance of VHMC on multi-modal distributions and real datasets and compare our method with the state-of-art algorithms. All our experiments are conducted on a standard computer with 4.0 Ghz Intel core i7 CPU. First, we introduce the performance index which will be used in the following parts.

\textbf{Effective sample size}. The variance of a Monte Carlo sampler is determined by its effective sample size (ESS) \citep{Neal2010} which is defined as:
\begin{equation}
\mathrm{ESS}=N/(1+2\times\sum_{s=1}^{\infty}\rho(s)),
\end{equation}
where $N$ represents the number of all the samples and $\rho(s)$ represents the $s-$step autocorrelation where autocorrelation is an index which considers the correlation between two samples. Let $X$ be a set of samples, and $t$ be the number of iteration ($t$ is an integer). Then $X_t$ is the sample at time $t$ of $X$. The autocorrelation between time $s$ and $t$ is defined as:
\begin{equation}
R(s,t)=\frac{E[(X_t-\mu_t)(X_s-\mu_s)]}{\sigma_t\sigma_s},
\end{equation}
where $E$ is the expected value operator. The correlation between two nearby samples can be measured with autocorrelation. The lower the value of autocorrelation is, the more independent the samples are.

\textbf{Maximum mean discrepancy}. The difference between samples drawn from two distributions can be measured as maximum mean discrepancy (MMD) \citep{gretton2012kernel} which takes the form as:
\begin{equation}
\begin{aligned}
MMD^{2}[X, Y]=&\frac{1}{M^2}\sum_{i,j=1}^{M}k(x_i,x_j)-\frac{2}{MN}\sum_{i,j=1}^{M,N}k(x_i,y_j)+\frac{1}{N^2}\sum_{i,j=1}^{N}k(y_i,y_j),
\end{aligned}
\end{equation}
where $M$ represents the sample number in $X$, $N$ represents the sample number in $Y$ and $k$ represents the kernel function. Through calculating the MMD value, we can analyze the convergence rate of the proposed methods.

\textbf{Relative error of mean}. It is a summary of the errors in approximating the expectation of variables across all dimensions \citep{ahn2013distributed}, which is computed as:
\begin{equation}
REM_t=\frac{\sum_{i=1}^{d}{\mid\bar{\theta_i^t}-\theta_i^*\mid}}{\sum{\mid \theta_i^*\mid}},
\end{equation}
where $\theta_i^*$ is the average of the $i$'th variable at time $t$, and $\theta_i^*$ is the actual mean value.

\subsection{Mixture of Isotropic Gaussians}
We conduct our first experiment on two multi-modal distributions where we consider two simple 2$D$ Gaussian mixture whose distribution is analytically available. First, we consider a Gaussian mixture distribution whose modes are close to each other and then we consider a Gaussian mixture whose modes are isolated and far away from each other. The distributions are given as follows:
$p(\textbf{x})=\frac{1}{2}\mathcal{N}(\textbf{x};\mu,\Sigma)+\frac{1}{2}(\textbf{x};-\mu,\Sigma)$ for $\sigma_x^{2}=\sigma_y^{2}=1$, $\rho_{xy}=0$, $\textbf{x}=(x.y)\in \mathbb{R}$ and $\mu=(2.5,-2.5)$ (modes are close to each other) or $\mu=(6.5,-6.5)$ (modes are far away from each other). The experiment setting is the same with \citet{Tripuraneni2017}. This multi-modal sampling problem is difficult for HMC, especially when the modes are far away from each other. The tremendous boundary gradient value in Hamiltonian dynamics forces HMC to fall into one of the two modes. Since the gradients in low probability region are large, the momentum will increasingly decrease until it changes its direction, which makes HMC sampler challenging to travel across these regions. The purpose of the experiments is to sample points which are independent identically distributed in these multi-modal distributions correctly.

In this experiment, we compare MHMC, HMC, MGHMC \citep{zhang2016towards} against VHMC. First, we compare the sample result of these methods intuitively. Then averaged autocorrelation and MMD are used to compare the performance of each method further. Each method is run for 10,000 iterations with 1,000 burn-in samples. The number of leap-frog steps is uniformly drawn from $(100-l,100+l)$ with $l=20$ which is suggested by \citet{Samuel2016}. We set step size $\epsilon=0.05$, friction coefficient $\gamma=0.5$ and the initiate position $\theta=(0,0)$. \citet{Tripuraneni2017} indicated that multi-modal problem is a challenge for HMC samplers. However, we find that HMC samplers are able to sample points from the multi-modal distribution especially when the modes are close to each other.

\begin{figure}[ht]

\begin{center}
    \subfigure[HMC-C ]
	{\includegraphics[width=0.3\columnwidth,height=4.4cm]{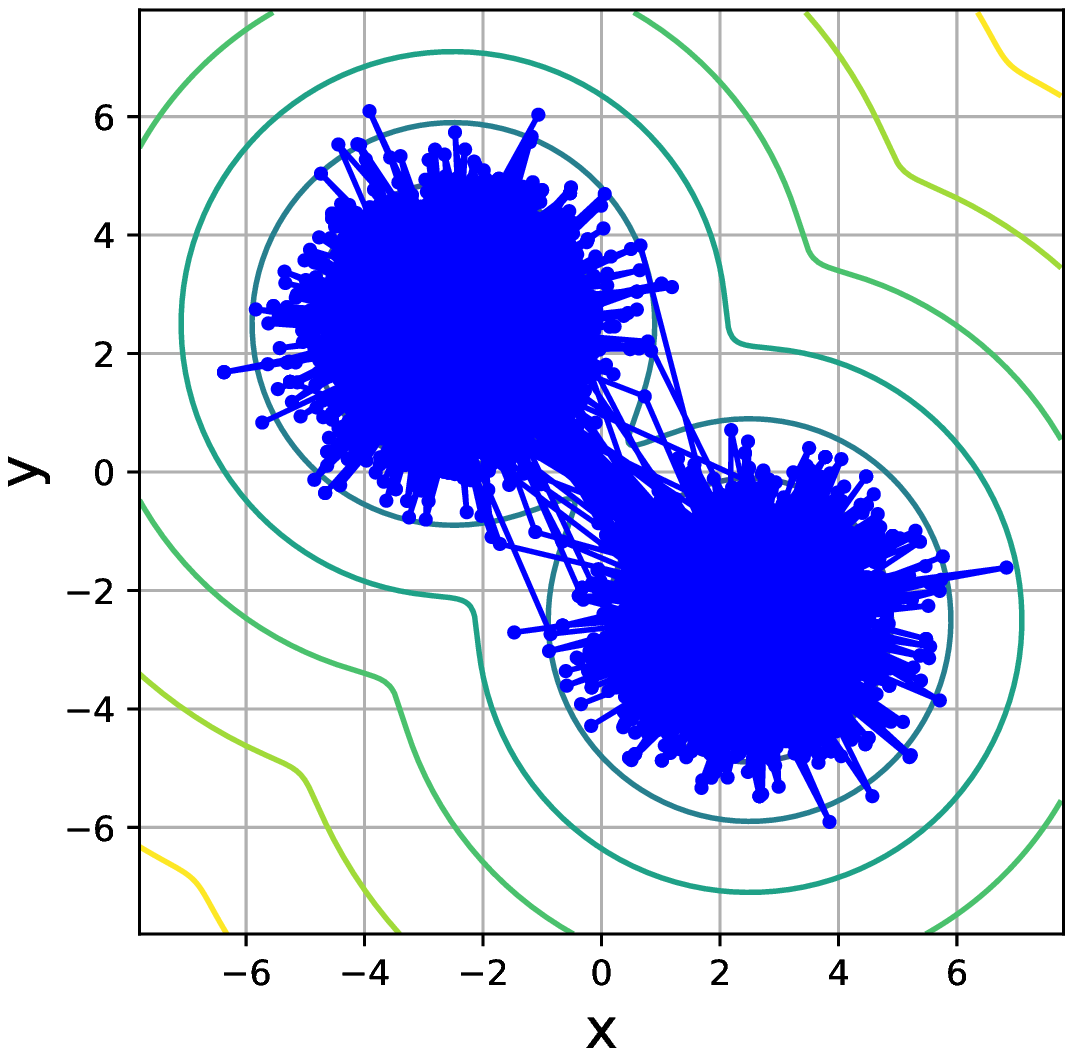}}
    \subfigure[MHMC-C ]
	{\includegraphics[width=0.3\columnwidth,height=4cm]{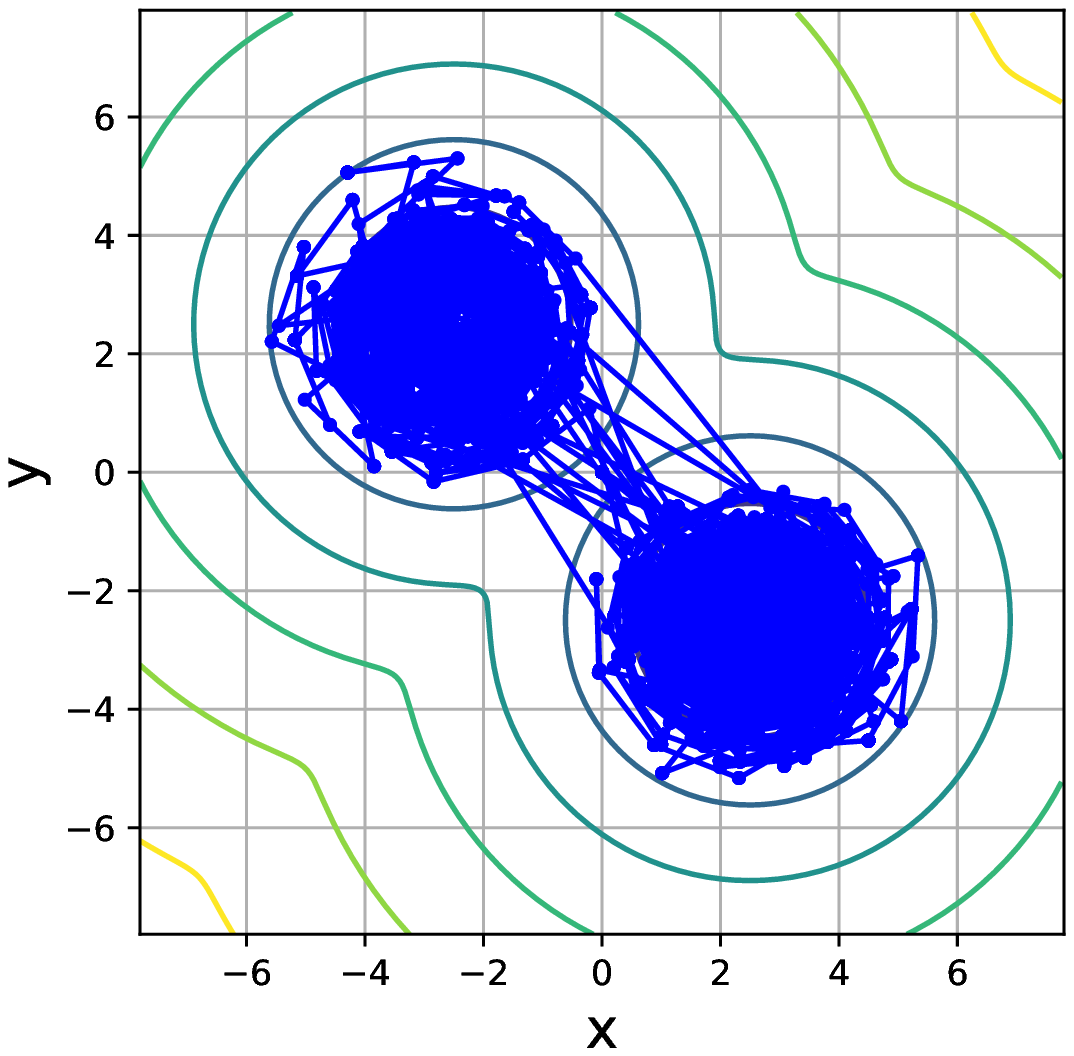}}
    \subfigure[VHMC-C ]
    {\includegraphics[width=0.3\columnwidth,height=4.4cm]{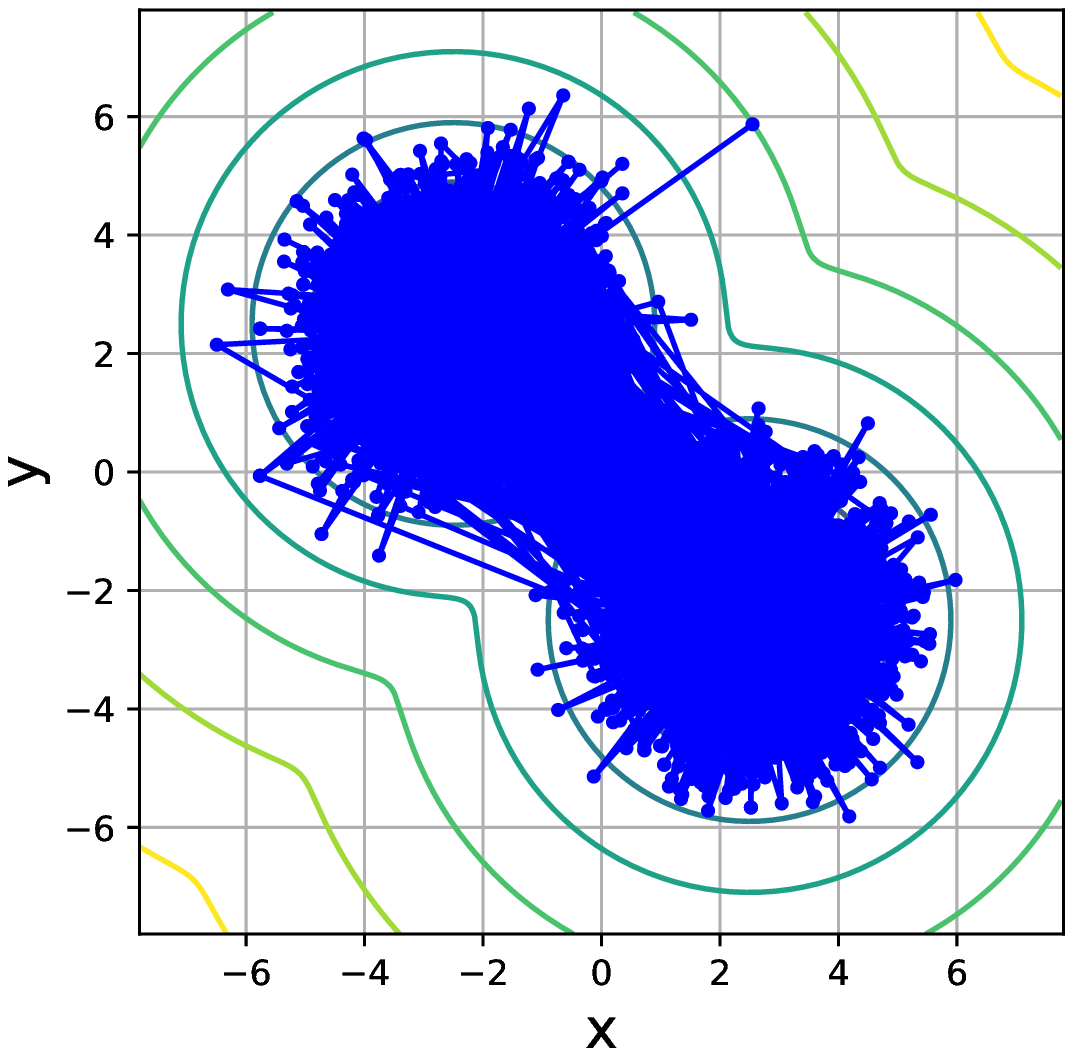}}
    \subfigure[HMC-F ]
    {\includegraphics[width=0.3\columnwidth,height=4.4cm]{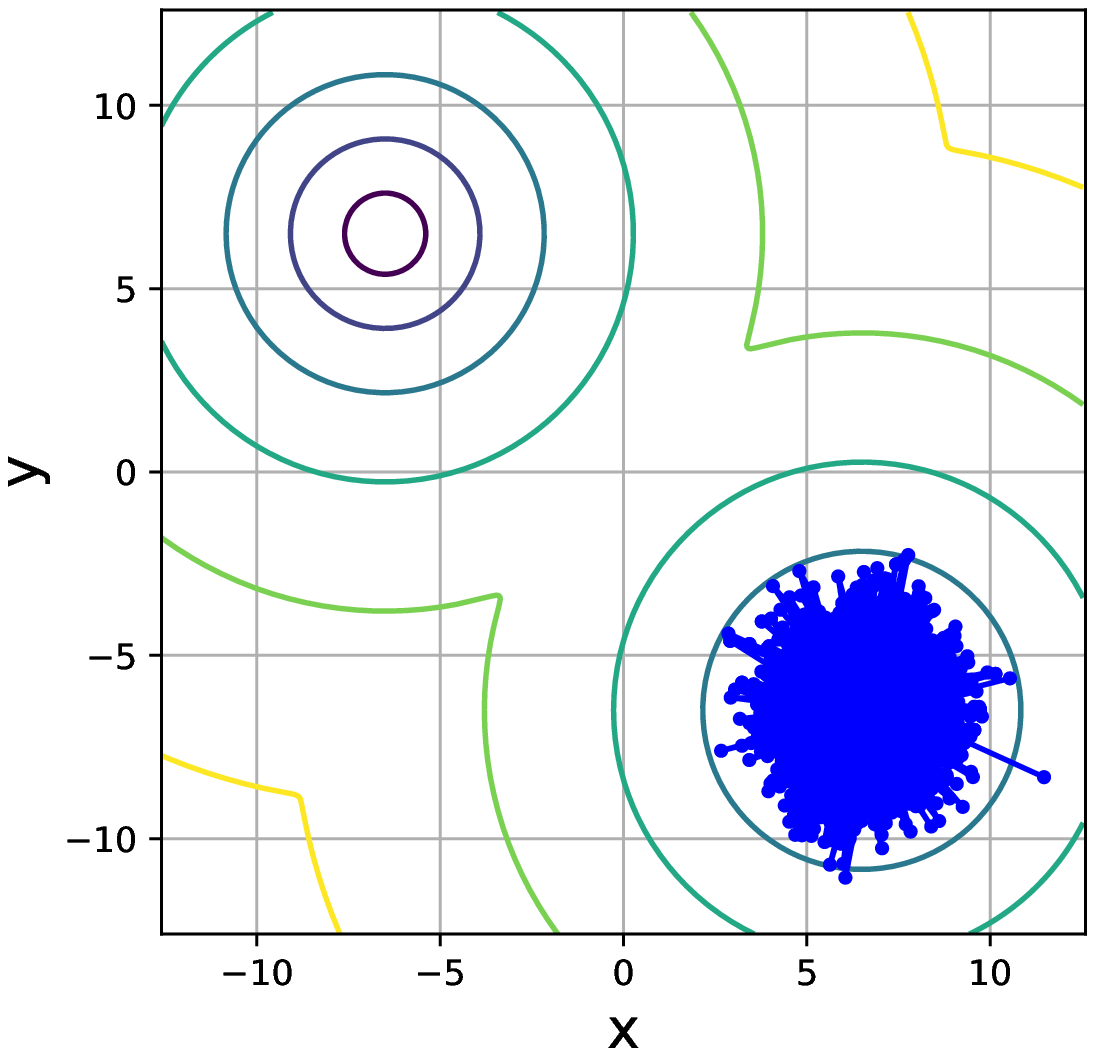}}
    \subfigure[MHMC-F ]
	{\includegraphics[width=0.3\columnwidth,height=4cm]{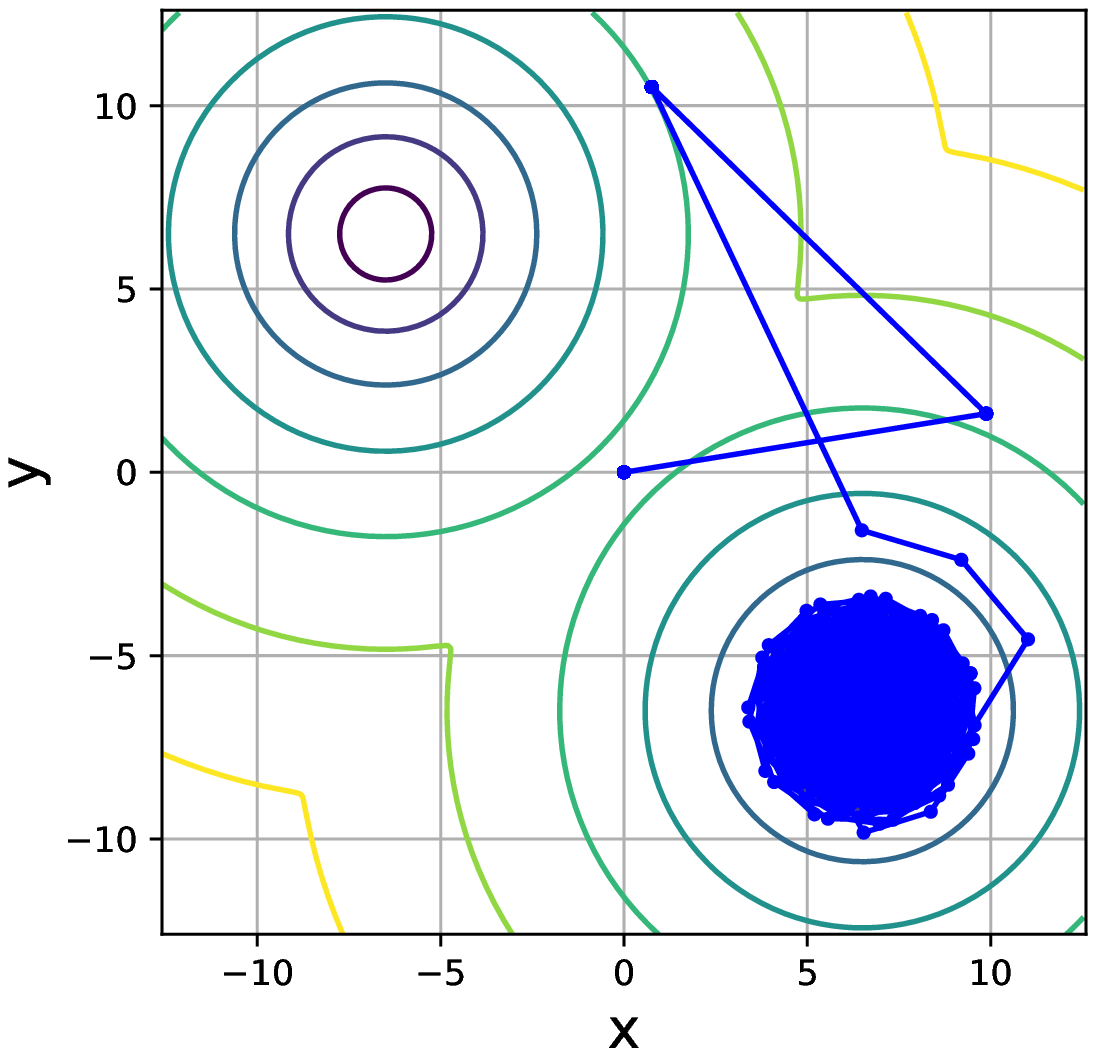}}
    \subfigure[VHMC-F ]
    {\includegraphics[width=0.3\columnwidth,height=4.4cm]{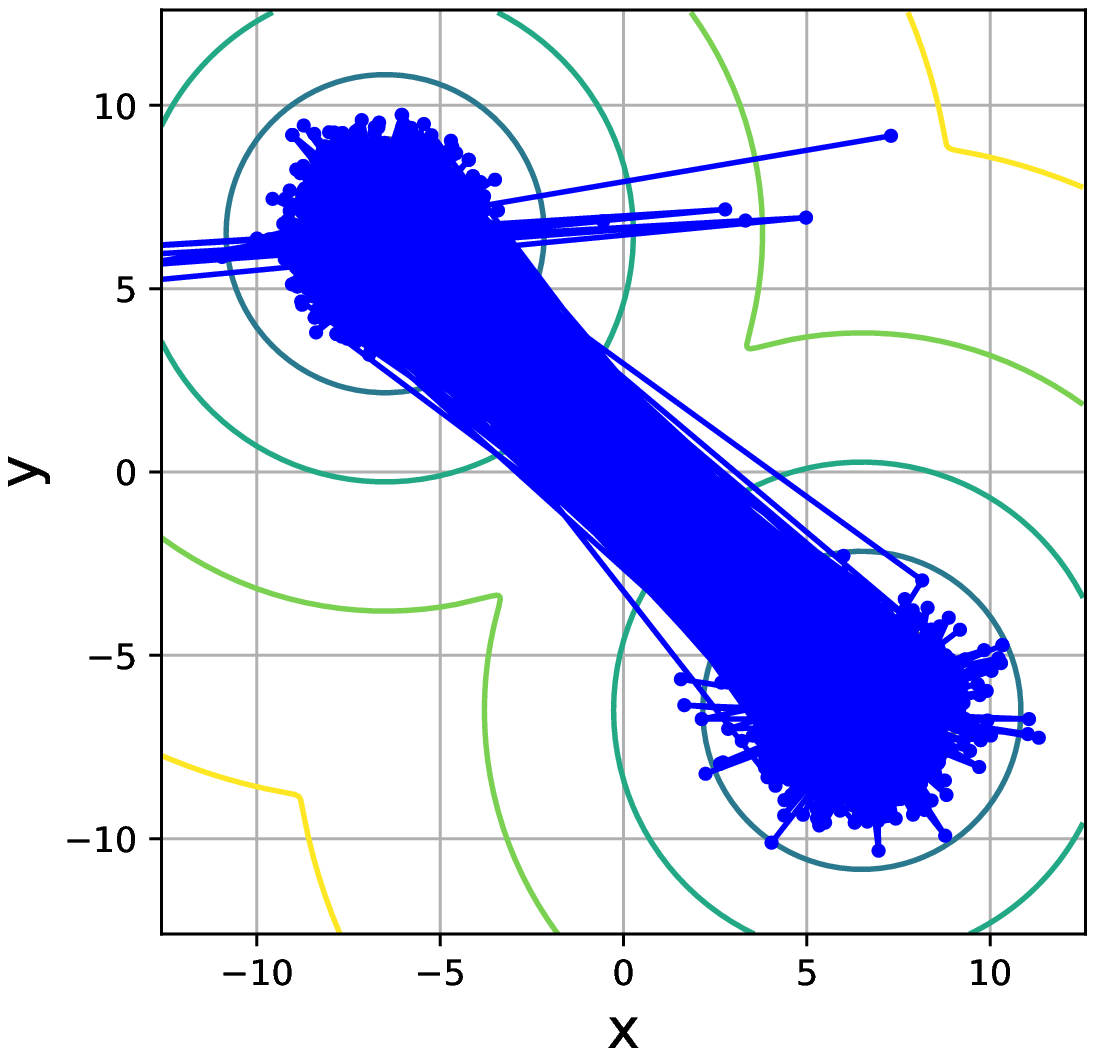}}
	\caption{Sampling experiment results on mixture Gaussian distributions. In the first line, HMC-C and MHMC-C and VHMC-C represent HMC, MHMC and VHMC sample from a Gaussian mixture, whose modes are close to each other and the mean value of each mode of the Gaussian mixture is $\mu_0=(2.5,-2.5)$ and $\mu_1=(-2.5,2.5)$ respectively. In the second line, HMC-F, MHMC-F and VHMC-F represent HMC, MHMC and VHMC sample from a Gaussian mixture, whose modes are far away from each other and the mean value of each mode of the Gaussian mixture is $\mu_0=(6.5,-6.5)$ and $\mu_1=(-6.5,6.5)$ respectively.}
    \label{guideandmix}
\end{center}
\end{figure}

Figure~\ref{guideandmix} clearly show that when $\mu=(2.5,-2.5)$, three methods can sample the multi-modal distribution. Nevertheless, there is some difference between them. HMC may sample from this mixture Gaussian distribution, but it hardly changes its sampling mode. MHMC sampler changes its sampling mode more frequently. While VHMC changes its mode much more frequent than MHMC. From the result, we can also conclude that when the modes are close to each other, HMC may sample this multi-modal distribution.

However, because HMC hardly changes its mode, it converges to the target distribution slowly, while VHMC changes its mode very frequently, which makes VHMC converge to the target distribution quickly. In order to compare the convergence rate and the independence of the samples with state-of-the-art sampling methods, we exploit MMD and autocorrelation to describe the performance when sampling the Gaussian mixture.

\begin{figure}[ht]
    \includegraphics[width=0.7\columnwidth,height=8cm]{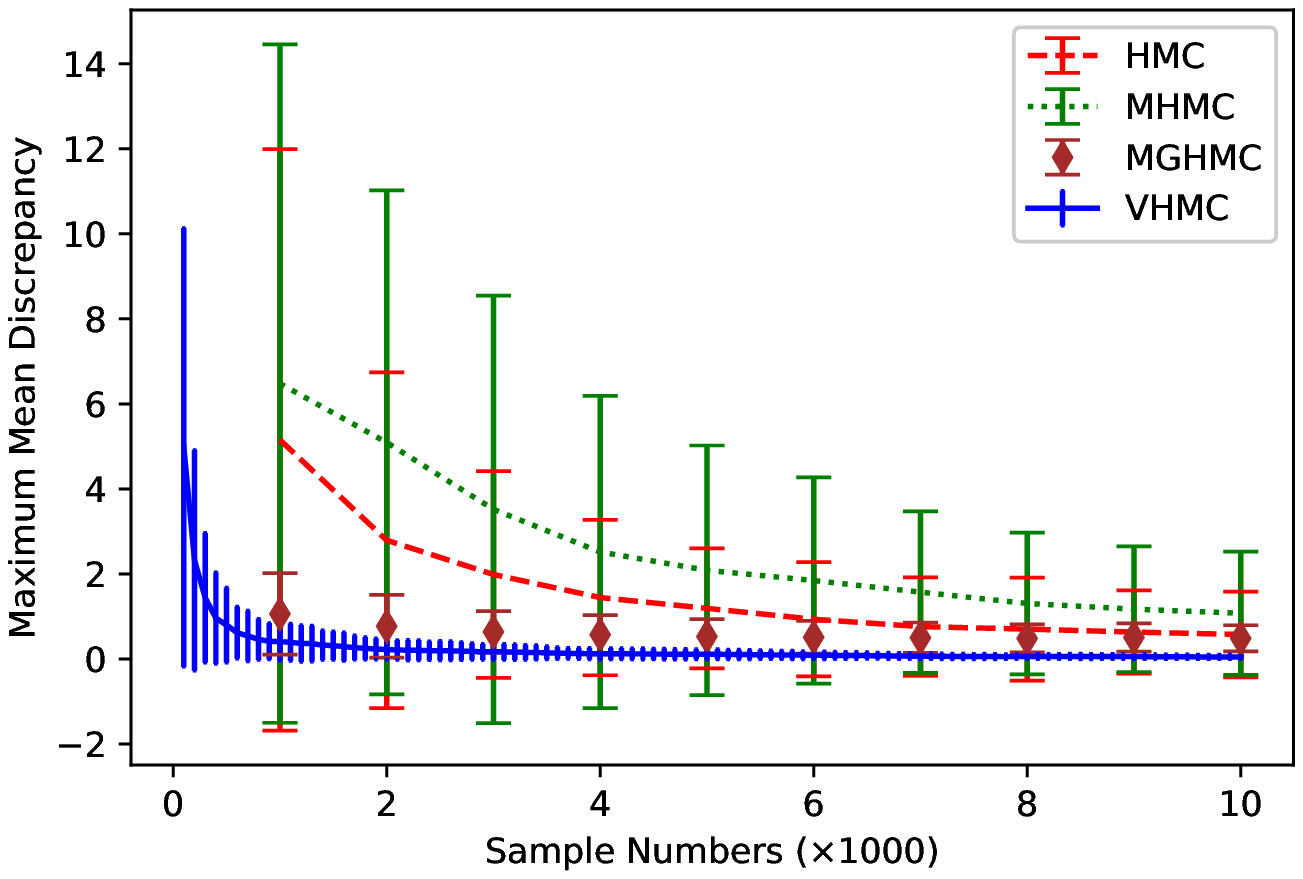}
    \includegraphics[width=0.7\columnwidth,height=8cm]{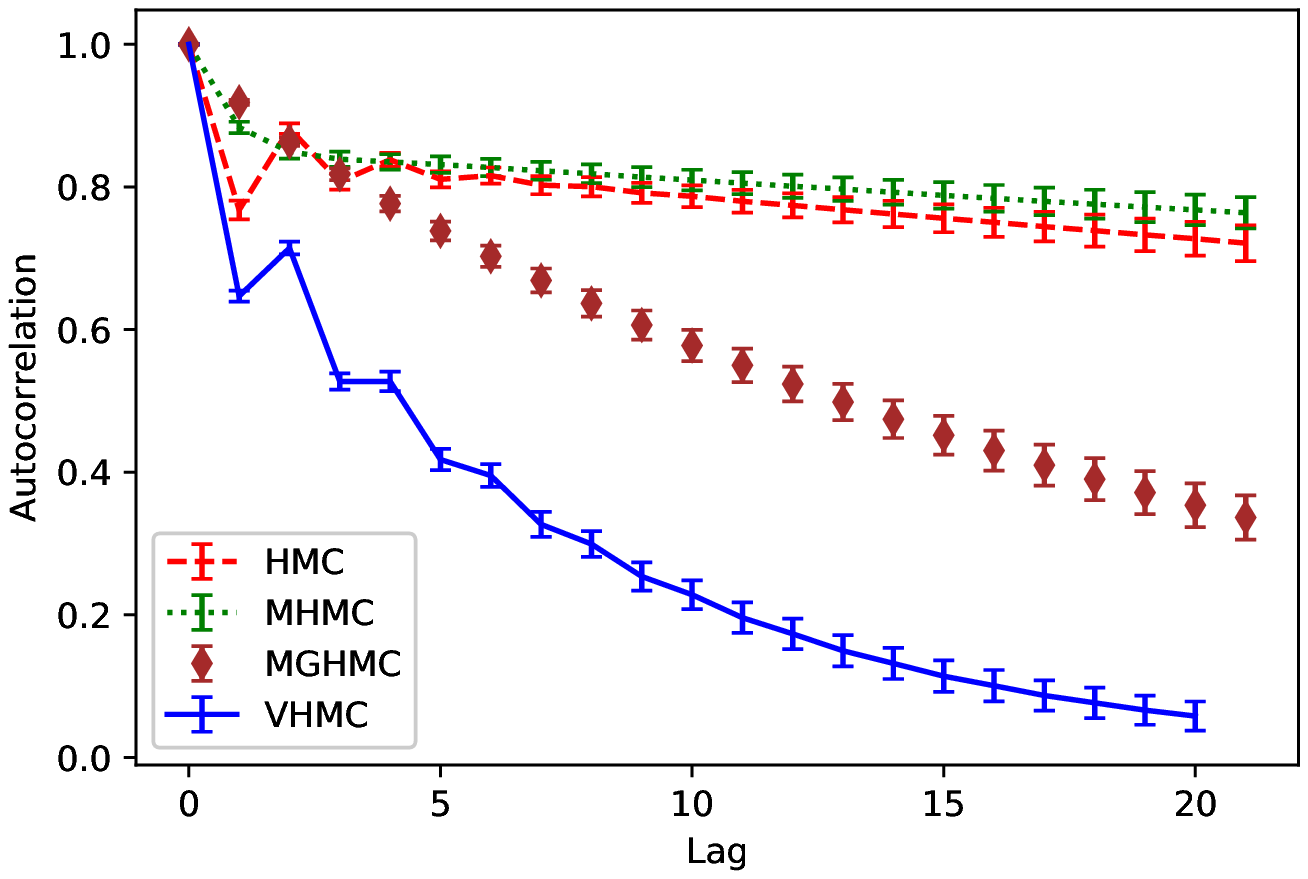}
    \centering
    \caption{The comparison of MMD and autocorrelation for four different methods. The upper row shows the relationship between MMD and lag of sample number and the bottom row demonstrates the relationship between autocorrelation and sample number.}
    \label{MMD-AUTOco}
\end{figure}

MMD between exact samples generated from the target density and samples generated from HMC, MHMC, MGHMC and VHMC chains was used to describe the convergence performance of the samplers. We use a quadratic kernel \citep{Karsten2006} $k(x,x^{'}=(1+\langle x,x^{'}\rangle)^2$ and averaged over 100 runs of the Markov chains. Figure~\ref{MMD-AUTOco} demonstrates that our method achieves the best performance in convergence rate and autocorrelation. Since our method converges to the target distribution quickly, we furthermore narrow the number of the first 500 samples.

We have already discussed the multi-modal distributions whose modes are close to each other and then we discuss the circumstance in which the modes are isolated and far away from each other. When $\mu$ in mixture Gaussian become larger, for instance, $\mu=(6.5,-6.5)$. The second row of Figure~\ref{guideandmix} shows that both HMC and MHMC can not sample from the target distribution. Nevertheless, our method still performs well. In Hamiltonian dynamics, there exists a significant force in this low probability regions which hinder samplers in jumping out of the current mode. VHMC takes advantages of the variational distribution to explore the phase space which results in excellent performance.

\begin{figure}[htbp]
\centering
\includegraphics[width=10cm,height=6cm]{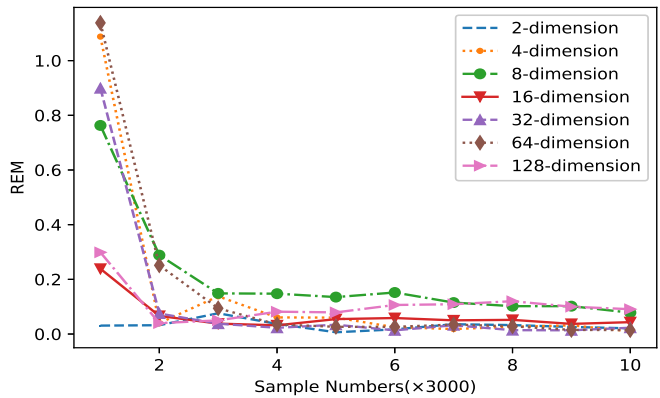}
  \caption{Relative error of mean on high dimensions.}
  \label{REM}
\end{figure}

To test the performance of the proposed method on high dimensional multi-modal distribution. We conduct our experiments on 2 to 128 dimensions. The target distribution is given as $p(\theta)=\frac{1}{(2\pi)^{\frac{n}{2}}}\biggl(0.7{\rm{exp}}(\frac{{-(x-\mu_0)^\top(x-\mu_0)}}{2})+0.3{\rm{exp}}(\frac{{-(x-\mu_1)^\top(x-\mu_1)}}{2})\biggr)$, where $\mu_0=(a_1,...,a_n), \mu_1=(b_1,...b_n), a_i=-1,b_i=1$ and $n$ equals dimensions. Figure~\ref{REM} shows that the proposed method has lower REM in high dimensions, which indicates that VHMC is able to sample from the high dimensional distant multi-modal distributions.

\begin{figure}[ht]
    \centering
    \includegraphics[width=.55\columnwidth,height=5cm]{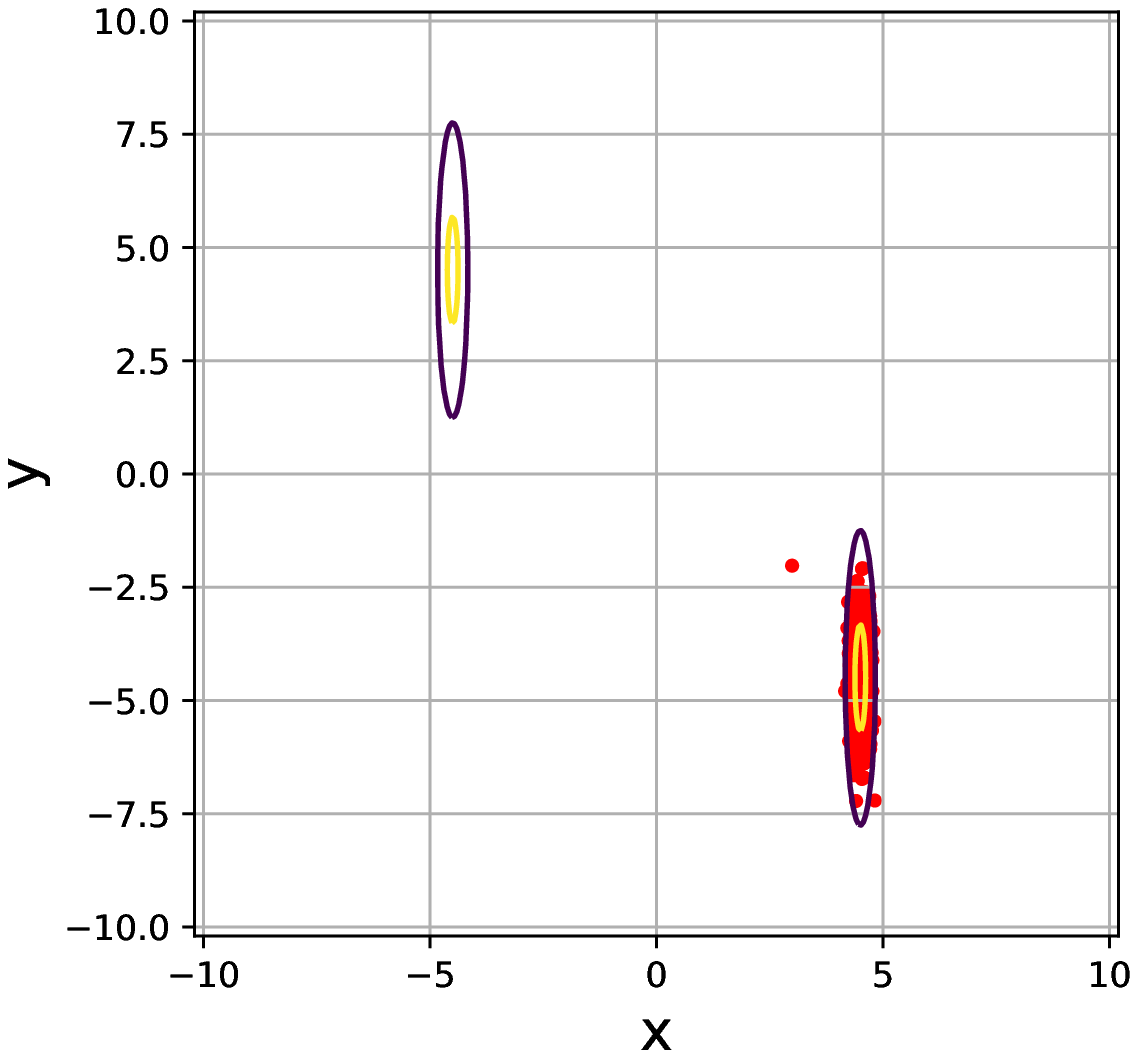}
    \includegraphics[width=.4\columnwidth,height=5cm]{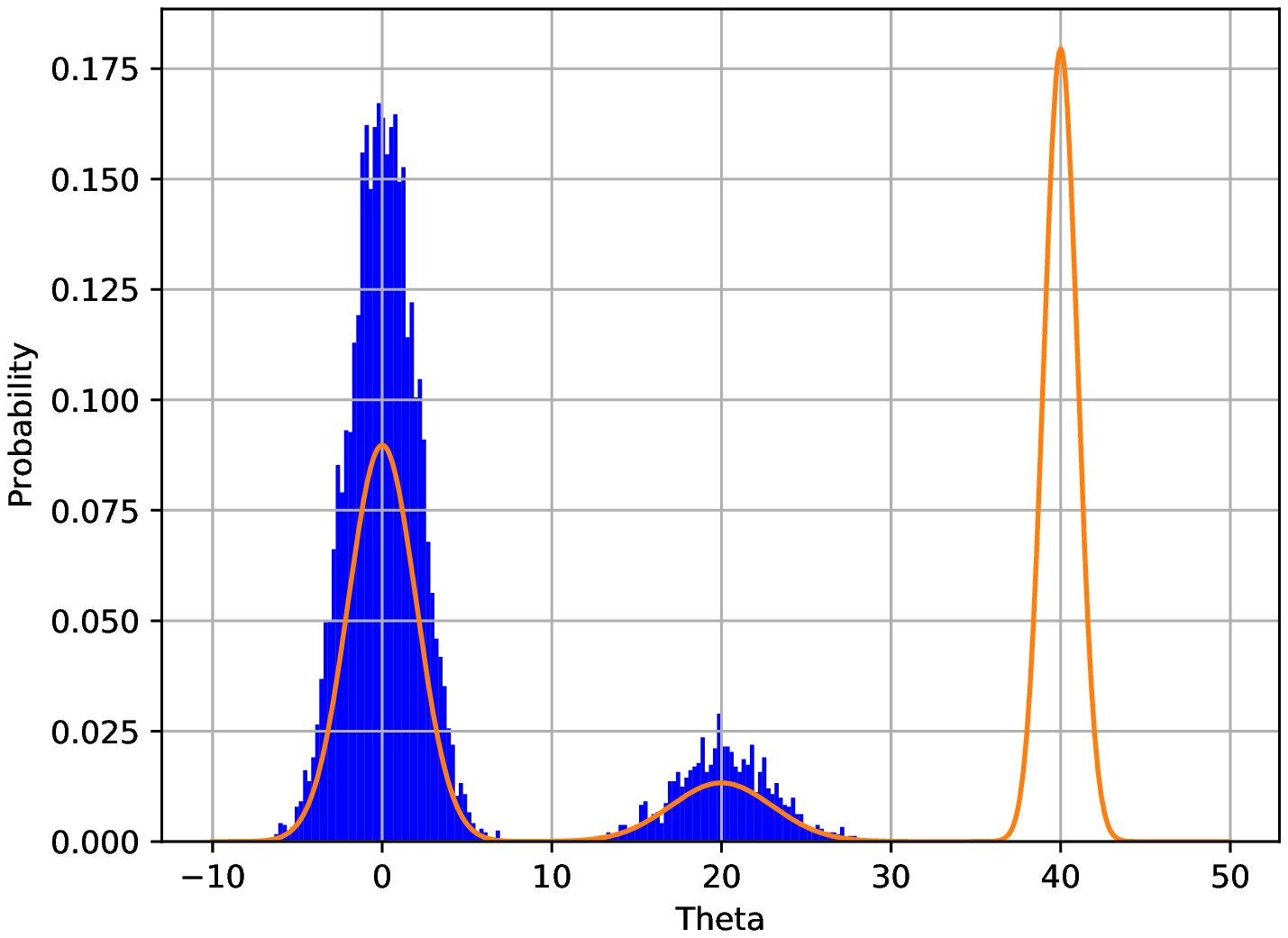}
    \includegraphics[width=.55\columnwidth,height=5cm]{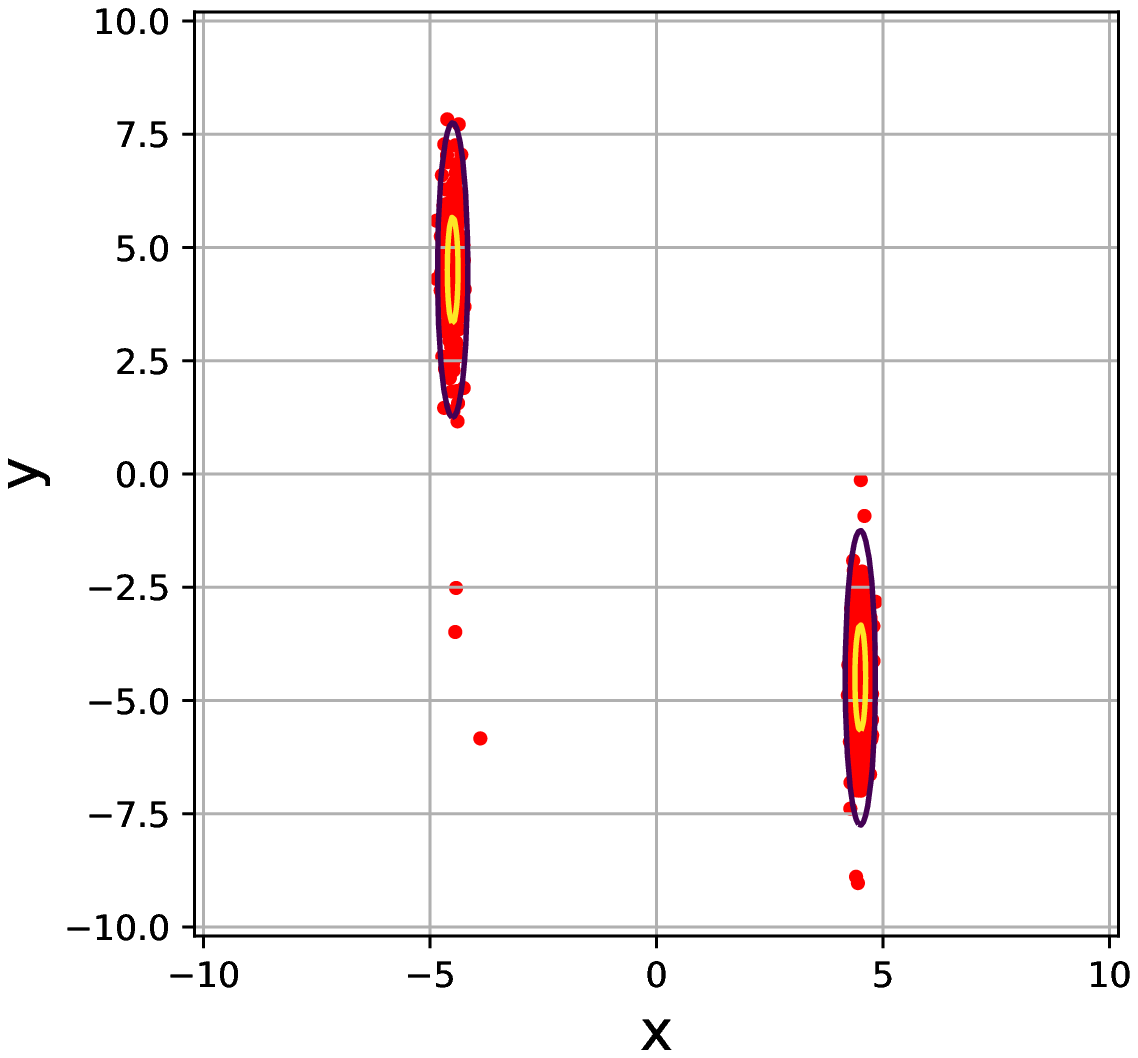}
    \includegraphics[width=.4\columnwidth,height=5cm]{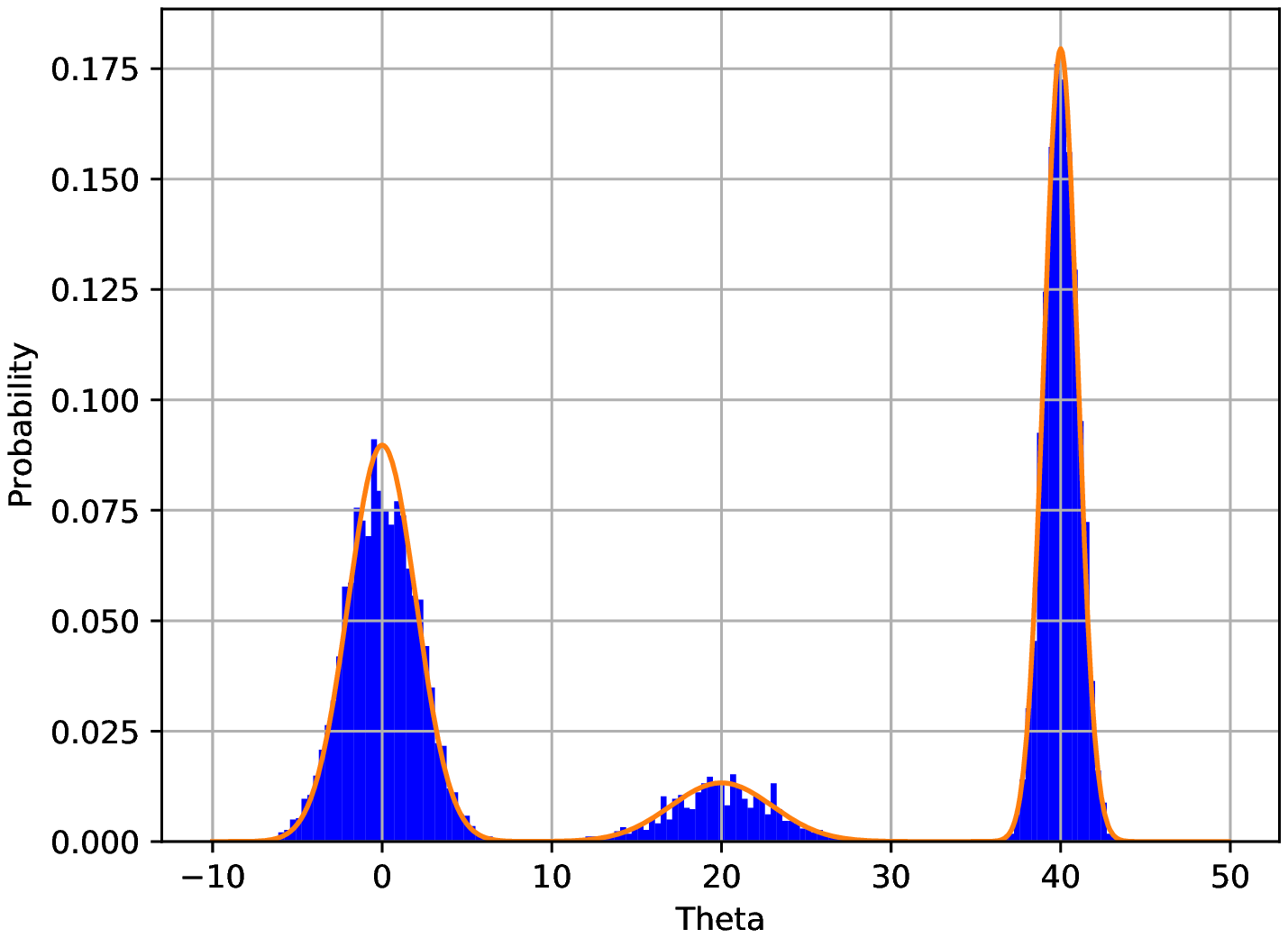}

    \caption{The performance of HMC and VHMC on the mixture of heterogeneous Gaussians. In the first column, we show the scatter diagram of HMC (upper) and VHMC (bottom). In the second column, we show the histgram of HMC (upper) and VHMC (bottom).}
    \label{VHMCDV}
\end{figure}

\subsection{Mixture of Heterogeneous Gaussians}
In the first experiment, we have already discussed the Gaussian mixture when the variance of the modes is the same. In practice, real data distributions often have different variances and probability of modes. In order to demonstrate the strong stability, we construct two new mixture Gaussian with different variances and probability of modes. The first one is given as follows:
$p(\theta)=\pi_1\mathcal{N}(\theta;\mu_1,\sigma_1)+\pi_2\mathcal{N}(\theta;\mu_2,\sigma_2)+\pi_3\mathcal{N}(\theta;\mu_3,\sigma_3).$ We set $\pi_1=0.1$, $\pi_2=0.8$, $\pi_3=0.1$, $\sigma_1=1$, $\sigma_2=3$, $\sigma_3=2$. The second one takes the form as: $p(\theta)=0.5\mathcal{N}(\theta;\mu_1,\sigma_1)+0.5\mathcal{N}(\theta;\mu_2,\sigma_2)$. Here we set $\sigma_x^{2}=0.01$, $\sigma_y^{2}=1$, $\rho_{xy}=0.0$. Similar to the previous experiment, our method runs 10000 iterations with 1,000 burn-in samples. Figure~\ref{VHMCDV} shows that VHMC has strong stability. Even when the variance becomes tiny, our method still shows the advanced performance. From the second column of Figure~\ref{VHMCDV} we can also observe that HMC sampler may sample multi-modal distribution especially when HMC sampler has chances to jump out of one mode. Although the distance of the left mode and the middle mode is the same with the distance of middle mode and right mode, the different variances force HMC sampler to sample from the left two modes.

\subsection{Bayesian Logistic Regression}
Logistic regression (LR) \citep{Freedman2009} is a traditional method for classification. We optimize the parameters by maximizing the logistic likelihood function. Exploiting the parameters, we can predict the class of the data.

To verify the performance on real datasets, we apply the proposed method to Bayesian logistic regression (BLR)\citep{mackay1992evidence} and our method is compared with logistic regression (LR), variational Bayesian logistic regression (VBLR) and HMC.

The likelihood function of a two-class classification problem can be defined as:
\begin{equation}
p({\rm{t}}|w)=\prod_{n=1}^{N}[1-y_n]^{1-t_n},
\end{equation}
where $t_n\in\{0,1\}$ and ${\rm t}=(t_1, ...,t_N)^\top$ and $y_n=p(\mathcal{C}_1|\phi_n)=\sigma(w^{\top}\phi)$. $t_n$ represents the label of the data and $y_n$ represents the predict value. We obtain the class of the data by means of integrating the logistic function on the posterior distribution.

We evaluate our methods on eight real-world datasets from UCI repository \citep{dataset2013}: Pima Indian (PI), Haberman (HA), Mammographic (MA), Blood (BL), Cryotherapy (CR), Immunotherapy (IM), Indian (IN), Dic (DI) using Bayesian logistic regression. The eight datasets are normalized to have zero mean value and unit variance. We give the Gaussian distribution $\mathcal{N}(0,100\textbf{I})$ as the prior distribution of the parameters.

In each experiment, we run $10000$ iterations with $2000$ burn-in samples. We draw leap-frog steps from a uniform distribution $\textbf{U}(80,120)$. We set step size $\epsilon=0.00045$ and mass matrix $m=3\textbf{I}$. On each dataset, we run 100 times to calculate the mean and the standard deviation.

Results in terms of the accurate rate of prediction and area under the ROC curve (AUC) \citep{Hanley1983} are summarized in Table~\ref{accuracy} and Table~\ref{ROC}. The results show that in these eight datasets, VHMC achieves better performance in classification accuracy rate and provide the similar performance with VBLR and better performance than HMC, which indicates that the method proposed in this paper can sample actual posterior distribution.

\begin{table}[t]
\caption{Classification accuracies for variational Bayesian logistic regression (VBLR), logistic regression (LR) HMC and VHMC on eight data sets.}
\label{accuracy}
\vskip 0.15in
\begin{center}
\begin{small}
\begin{sc}
\begin{tabular}{lcccr}
\toprule
DATA & LR & VBLR & HMC & VHMC \\
\midrule
HA &67.9$\pm$0.6& 67.7$\pm$0.5&67.7$\pm$0.4&68.2$\pm$0.5\\
PI &82.5$\pm$0.2& 82.7$\pm$0.2&82.4$\pm$0.3&83.1$\pm$0.2\\
MA &89.8$\pm$0.1& 89.9$\pm$0.1&89.9$\pm$0.2&89.9$\pm$0.2\\
BL &75.1$\pm$0.3& 75.3$\pm$0.3&71.2$\pm$0.6&75.4$\pm$0.3\\
CR &95.4$\pm$0.1& 95.6$\pm$0.1&92.1$\pm$0.2&95.7$\pm$0.1\\
IM &77.5$\pm$0.4& 77.5$\pm$0.4&77.4$\pm$0.4&77.6$\pm$0.4\\
IN &75.5$\pm$0.2& 75.8$\pm$0.2&75.3$\pm$0.3&75.9$\pm$0.3\\
DI &82.6$\pm$0.2& 82.5$\pm$0.2&81.8$\pm$0.3&82.5$\pm$0.2\\
\toprule
RANK & 2.75 & 2.125 & 3.5 & 1.25 \\
PVALUE & 0.012 & 0.041 & 0.056 & /\\
\bottomrule
\end{tabular}
\end{sc}
\end{small}
\end{center}
\vskip -0.1in
\end{table}

\begin{table}[t]
\caption{Area under the ROC curve for VBLR, LR, HMC and VHMC on eight data sets.}
\label{ROC}
\vskip 0.15in
\begin{center}
\begin{small}
\begin{sc}
\begin{tabular}{lcccr}
\toprule
DATA & LR & VBLR & HMC & VHMC \\
\midrule
HA &73.8$\pm$0.4& 74.6$\pm$0.4&74.6$\pm$0.2&74.6$\pm$0.4\\
PI &77.0$\pm$0.2& 77.1$\pm$0.2&77.2$\pm$0.3&77.6$\pm$0.2\\
MA &82.6$\pm$0.2& 82.7$\pm$0.1&82.5$\pm$0.2&82.8$\pm$0.1\\
BL &77.1$\pm$0.2& 77.1$\pm$0.2&74.4$\pm$0.5&77.2$\pm$0.2\\
CR &86.9$\pm$0.2& 87.2$\pm$0.2&84.2$\pm$0.3&87.3$\pm$0.2\\
IM &84.7$\pm$0.2& 84.8$\pm$0.2&83.8$\pm$0.2&84.9$\pm$0.2\\
IN &72.7$\pm$0.2& 72.8$\pm$0.2&71.3$\pm$0.3&72.9$\pm$0.2\\
DI &74.4$\pm$0.2& 74.6$\pm$0.2&74.0$\pm$0.2&74.7$\pm$0.2\\
\toprule
RANK & 3.125 & 2.125 & 3.375 & 1.125 \\
PVALUE & 0.004 & 0.036 & 0.018 & /\\
\bottomrule
\end{tabular}
\end{sc}
\end{small}
\end{center}
\vskip -0.1in
\end{table}

\section{Conclusion}
In this study, we presented VHMC, a novel sampling algorithm that aims to sample from the distant multi-modal distributions. Langevin dynamics and equipotential conversion are added in the proposed method to accelerate the convergence rate and reduce the autocorrelation of the samples. We exploit the information of the variational distribution of the target distribution to make effective distant multi-modal sampling available. Formal theoretical analysis is provided which demonstrated that VHMC could converge to the target distribution. Our findings are supported by synthetic and real data experiments which showed that VHMC brings multiple benefits, such as providing superior performance in multi-modal sampling and lower autocorrelation. In the future, we plan to apply stochastic gradient \citep{chen2014stochastic} to our method for scalable MCMC.

\acks{This work is supported by the National Natural Science Foundation of China under Project 61673179, and Shanghai Knowledge Service Platform Project (No. ZF1213). }


\bibliography{sample}

\end{document}